\documentclass{article}

\usepackage{arxiv}

\usepackage[utf8]{inputenc} 
\usepackage[T1]{fontenc}    
\usepackage{hyperref}       
\usepackage{url}            
\usepackage{booktabs}       
\usepackage{amsfonts}       
\usepackage{nicefrac}       
\usepackage{microtype}      
\usepackage{lipsum}		
\usepackage{graphicx}
\usepackage{natbib}
\usepackage{doi}
\usepackage{amsmath}
\usepackage{tikz}
\usepackage{geometry}
\usepackage{tikz}

\usetikzlibrary{mindmap, shadows, positioning, backgrounds}
\usetikzlibrary{positioning, shapes, arrows.meta}
\usepackage{amsmath, amssymb}

\title{UAVs Meet Agentic AI: A Multidomain Survey of Autonomous Aerial Intelligence and Agentic UAVs}

\author{
  \href{https://orcid.org/0000-0002-5417-6744}{\includegraphics[scale=0.06]{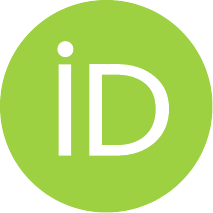}\hspace{1mm}Ranjan Sapkota\textsuperscript{1}} \quad
  Konstantinos I. Roumeliotis\textsuperscript{2} \quad
  \href{https://orcid.org/0000-0001-5337-4848}{\includegraphics[scale=0.06]{orcid.pdf}\hspace{1mm}Manoj Karkee\textsuperscript{1}} \\
  \textsuperscript{1}Cornell University, Department of Biological and Environmental Engineering,  Ithaca, NY 14853, USA \\
  \texttt{rs2672@cornell.edu, mk2684@cornell.edu} \\
  \textsuperscript{2}Department of Informatics and Telecommunications, University of the Peloponnese, Tripoli, Greece \\
}

\renewcommand{\shorttitle}{UAVs Meet Agentic AI}

\hypersetup{
pdftitle={A template for the arxiv style},
pdfsubject={q-bio.NC, q-bio.QM},
pdfauthor={David S.~Hippocampus, Elias D.~Striatum},
pdfkeywords={First keyword, Second keyword, More},
}

\begin{document}
\maketitle

\begin{abstract}
Agentic UAVs represent a new frontier in autonomous aerial intelligence, integrating perception, decision-making, memory, and collaborative planning to operate adaptively in complex, real-world environments. Driven by recent advances in Agentic AI, these systems surpass traditional UAVs by exhibiting goal-driven behavior, contextual reasoning, and interactive autonomy. We provide a comprehensive foundation for understanding the architectural components and enabling technologies that distinguish Agentic UAVs from traditional autonomous UAVs. Furthermore, a detailed comparative analysis highlights advancements in autonomy, learning, and mission flexibility. This study explores seven high-impact application domains \textit{precision agriculture},  \textit{construction \& mining}, \textit{disaster response}, \textit{environmental monitoring}, \textit{infrastructure inspection}, \textit{logistics}, \textit{security}, and  \textit{wildlife conservation}, illustrating the broad societal value of agentic aerial intelligence. Furthermore, we identify key challenges in technical constraints, regulatory limitations, and data-model reliability, and we present emerging solutions across hardware innovation, learning architectures, and human-AI interaction. Finally, a future roadmap is proposed, outlining pathways toward self-evolving aerial ecosystems, system-level collaboration, and sustainable, equitable deployments. This survey establishes a foundational framework for the future development, deployment, and governance of agentic aerial systems across diverse societal and industrial domains.
\end{abstract}

\section{Introduction}
\subsection{Motivation and Scope}

Unmanned Aerial Vehicles (UAVs) have rapidly evolved from remote-controlled platforms for aerial imaging \cite{pajares2015overview, kim2019unmanned} into sophisticated autonomous agents and multi-agent systems(MAS) capable of performing complex tasks across a wide range of domains \cite{fotohi2020agent, vallejo2020multi, zhao2024graph}. This evolution has been accelerated by advances in artificial intelligence (AI), particularly through the integration of cognitive architectures \cite{chong2007integrated, thorisson2012cognitive, lieto2018role} that endow UAVs with planning, reasoning, and adaptive decision-making abilities \cite{javaid2025explainable}. The convergence of UAV technologies with agentic AI the paradigm in which systems exhibit autonomous, goal-directed, and context-aware behavior marks a fundamental shift in how aerial systems are designed, deployed, and interpreted \cite{kourav2025artificial}. This shift enables UAVs to operate not merely as reactive tools but as proactive entities capable of engaging in mission planning, environmental interpretation, and collaborative action with minimal human oversight \cite{chitra2025artificial}.

Agentic AI introduces capabilities such as semantic perception, affordance reasoning, and reflective planning, allowing UAVs to dynamically respond to environmental stimuli, learn from prior experiences, and optimize mission outcomes in real time \cite{sapkota2025ai}. Unlike traditional automation, which is limited to predefined tasks, agentic UAVs function as intelligent agents that can decompose goals, resolve uncertainty, and adjust behavior based on both internal objectives and external constraints \cite{sapkota2025ai, raza2025trism}. This evolution is critical to addressing real-world challenges in complex and unstructured environments ranging from disaster response and infrastructure inspection to wildlife monitoring and precision agriculture.

The motivation for this review stems from the growing deployment of UAVs in domains where autonomy, resilience, and interpretability are not optional but essential. Across sectors such as agriculture, logistics, environmental monitoring, and public safety, the operational requirements for UAVs are becoming increasingly dynamic. For instance, a UAV deployed for post-disaster search-and-rescue must detect structural hazards, locate survivors, and coordinate with other robotic assets all while operating in communication-constrained and GPS-denied environments. Similarly, in precision agriculture, UAVs are now expected to perform high-resolution crop diagnostics, adaptive spraying, and real-time interaction with other smart systems based on semantic maps and agronomic data.

Despite these capabilities, there remains a lack of unified understanding and structured synthesis across domains on how agentic principles are implemented and leveraged in UAV systems. Most existing reviews are domain-specific, focusing on applications in agriculture or defense, and often emphasize the hardware or algorithmic components in isolation. A multidomain synthesis is essential to uncover shared architectural patterns, transferable AI models, and cross-sectoral deployment strategies that define the current trajectory of autonomous aerial intelligence.

This review aims to fill that gap by systematically examining the role of agentic AI in transforming UAVs into autonomous, interactive, and collaborative agents across multiple real-world domains. It investigates how autonomy is achieved through layered system architectures, how decision-making processes are grounded in multimodal perception and learning, and how UAVs are increasingly integrated into broader intelligent ecosystems. By analyzing the state-of-the-art across seven to eight key sectors, this survey not only provides a technical foundation for future research but also highlights the socio-technical implications of agentic UAV systems in public, industrial, and environmental contexts. In doing so, the study establishes a comprehensive framework to understand the cross-domain impact of agentic UAVs and outlines the emerging frontiers of their development and deployment.

\subsection{Defining Agentic UAVs}

Agentic UAVs represent a new class of autonomous aerial systems distinguished by their cognitive capabilities, contextual adaptability, and goal-directed behavior. Unlike conventional UAVs that operate based on predefined instructions \cite{sautenkov2025uav} or rule-based automation \cite{sieber2024rule, natarajan2025artificial}, agentic UAVs function as intelligent systems capable of perceiving their environment, making complex decisions, and executing actions that align with mission objectives in dynamic settings. The term "agentic" originates from cognitive science and denotes an entity that exhibits autonomy, intentionality, self-regulation, and adaptability traits now being engineered into next-generation UAVs through AI integration \cite{sapkota2025ai}.

At the core of an agentic UAV is a layered architecture that mirrors the sense-think-act loop found in biological agents \cite{chitra2025artificial}. The **perception layer** utilizes multimodal sensors such as RGB, thermal, LiDAR, hyperspectral, and environmental probes to acquire rich situational awareness. These inputs are interpreted through onboard or edge-optimized AI models, enabling the UAV to semantically understand objects, terrain, anomalies, and mission-critical features in real time. For instance, visual data may be fused with weather sensor inputs to dynamically update flight paths in response to wind gusts or thermal conditions \cite{tian2025uavs, zadeh2025conceptual}.

Building upon perception, the **cognition layer** embodies the decision-making core of the agentic UAV \cite{chitra2025artificial, sautenkov2025uav}. This layer includes modules for reasoning, task decomposition, affordance learning, and planning under uncertainty. Techniques such as reinforcement learning, transformer-based attention, and probabilistic modeling are commonly used to create adaptive control policies \cite{essaky2025arresvg}. This enables UAVs to make context-sensitive decisions such as choosing between multiple intervention strategies during a crop disease outbreak or prioritizing a high-risk zone during emergency reconnaissance \cite{liu2025review, natarajan2025artificial, hao2025uav}.

The **planning and control layer** operationalizes these decisions into concrete trajectories and actuation commands \cite{emami2024age, tang2025dnn}. Advanced path planning algorithms allow UAVs to avoid obstacles, reconfigure missions on-the-fly, and coordinate with other agents \cite{wenlong2025advances}. Importantly, agentic UAVs can also exhibit *reflective control*, where past experiences are used to inform current planning decisions, leading to continual self-improvement and mission robustness \cite{al2024progress, chitra2025artificial}. These capabilities are not isolated but integrated within a communication layer that enables interaction with humans, cloud platforms, or other autonomous systems. Through natural language interfaces, shared semantic maps, and synchronized mission logs, agentic UAVs become part of an intelligent, interoperable ecosystem.

\subsection{Survey Objectives and Contributions}

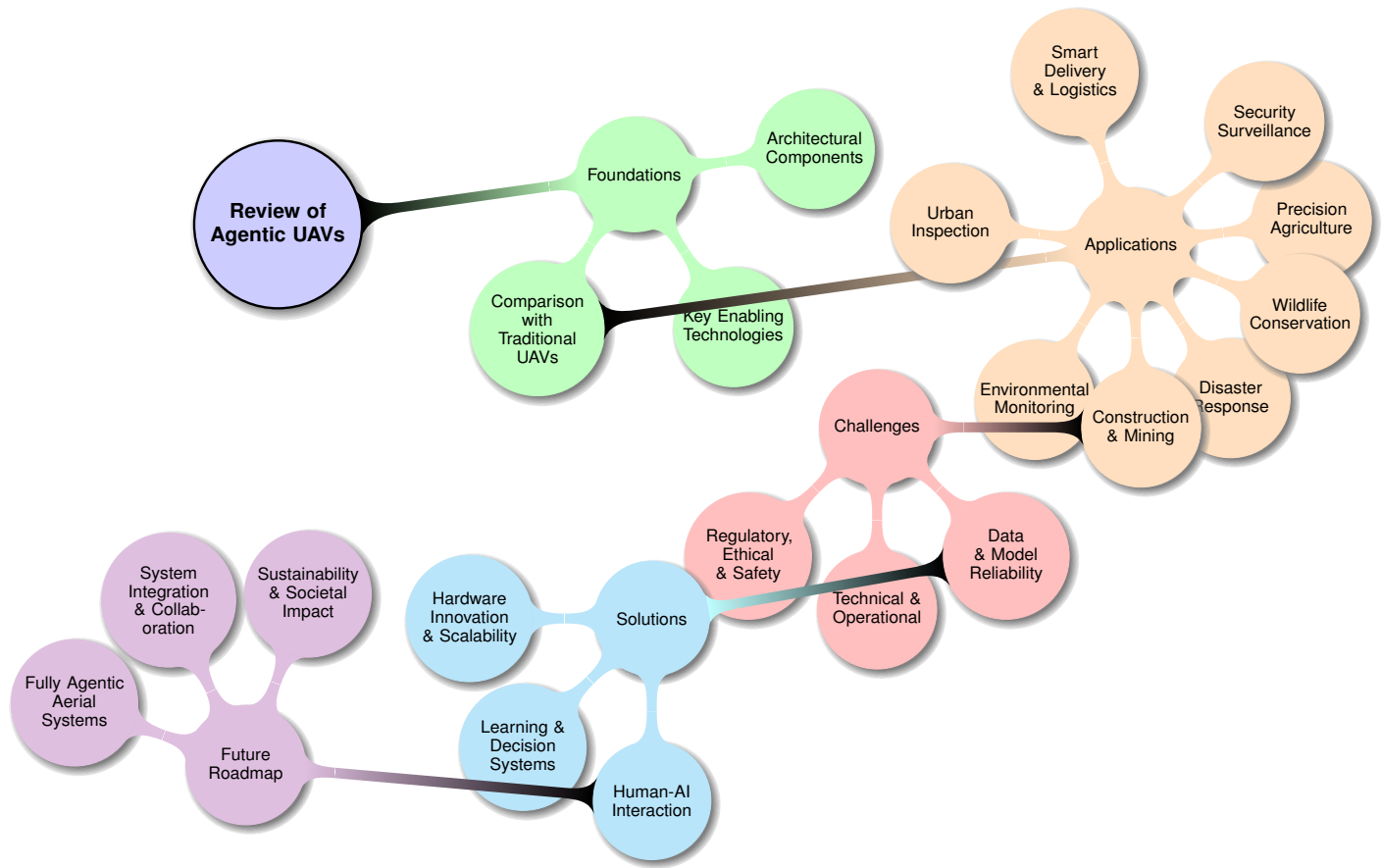
\begin{figure*}[ht]
\centering
\begin{tikzpicture}[
    scale=0.65,
    transform shape,
    mindmap,
    every node/.style={concept, circular drop shadow, font=\sffamily, text=black},
    root concept/.append style={
        font=\large\bfseries\sffamily,
        minimum size=3.5cm,
        text width=3 cm,
        fill=blue!20,
        line width=0.9pt,
        text=black
    },
    level 1 concept/.append style={
        sibling angle=75,
        level distance=7.5cm,
        text width=2.3cm,
        font=\scriptsize\sffamily,
        line width=0.5pt,
        fill=white
    },
    level 2 concept/.append style={
        sibling angle=65,
        level distance=3.8cm,
        text width=2.3cm,
        font=\tiny\sffamily,
        line width=0.3pt,
        fill=white
    }
]

\node [root concept] {Review of \\ Agentic UAVs}
  [clockwise from=8]

  child[concept color=green!25] {
    node {Foundations}
    child { node {Architectural \\ Components} }
    child { node {Key Enabling \\ Technologies} }
    child { node {Comparison with \\ Traditional UAVs} }
  }

  child[concept color=orange!25, level distance=12.5cm] {
    node {Applications}
    child { node {Precision \\ Agriculture} }
    child { node {Disaster Response \\} }
    child { node {Environmental \\ Monitoring} }
    child { node {Urban \\ Inspection} }
    child { node {Smart Delivery \\ \& Logistics} }
    child { node {Security \\ Surveillance} }
    child { node {Wildlife \\ Conservation} }
    child { node {Construction \\ \& Mining} }
  }

child[concept color=red!25, grow=180, level distance=5.5cm] {
    node {Challenges}
    child[grow=270] { node {Technical \& \\ Operational} }
    child[grow=225] { node {Regulatory, Ethical \\ \& Safety} }
    child[grow=315] { node {Data \\ \& Model \\ Reliability} }
}

child[concept color=cyan!25, grow=190] {
    node {Solutions}
    child[grow=180] { node {Hardware Innovation \\ \& Scalability} }
    child[grow=225] { node {Learning \& \\ Decision Systems} }
    child[grow=270] { node {Human-AI \\ Interaction} }
}

child[concept color=violet!25, grow=175, level distance=8.5cm] {
    node {Future Roadmap}
    child[grow=160] { node {Fully Agentic \\ Aerial Systems} }
    child[grow=115] { node {System Integration \\ \& Collaboration} }
    child[grow=70]  { node {Sustainability \\ \& Societal Impact} }
}
;

\end{tikzpicture}
\caption{Mind map of this review outlining the scientific foundation, application domains, technical challenges, emerging opportunities, and future roadmap of Agentic UAVs. The five main branches reflect the sequential structure of the paper from architectural components and multidomain use cases to system-level evolution and governance.}
\label{fig:agentic_uav_mindmap}
\end{figure*}

The primary objective of this review (\textbf{Figure \ref{fig:agentic_uav_mindmap}}) is to provide a structured and comprehensive synthesis of the emerging field of agentic UAVs across multiple real-world application domains. 

While existing reviews have addressed UAVs in isolated sectors such as agriculture, defense, or environmental monitoring there remains a significant gap in cross-domain analysis of how autonomy and cognitive intelligence are transforming UAV capabilities. This survey addresses that gap by introducing a unifying framework that defines the architectural, functional, and operational characteristics of agentic UAVs and explores their implementation in a variety of sectors that demand real-time decision-making, situational adaptability, and autonomous mission execution.

Specifically, the contributions of this review are threefold. First, it formulates a clear and actionable definition of agentic UAVs grounded in the principles of autonomous reasoning, multimodal perception, and reflective control. This helps distinguish agentic UAVs from traditional automated platforms and provides a baseline for evaluating the state-of-the-art.

Second, the review conducts an in-depth literature synthesis across seven to eight critical domains including precision agriculture, disaster response, infrastructure inspection, environmental surveillance, logistics, security, and ecological monitoring. For each domain, it highlights key technological enablers, representative use cases, and core research challenges, allowing readers to identify patterns and divergences in system design and deployment strategies.

Third, it outlines a future roadmap for research and development, identifying cross-cutting challenges related to power efficiency, onboard intelligence, swarm coordination, regulatory barriers, and data governance. This forward-looking perspective offers practical insights for researchers, developers, and policymakers aiming to build, adopt, or regulate intelligent aerial systems.

Through this multidomain analysis, the review aims to establish a foundational reference for the rapidly evolving field of agentic UAVs and catalyze interdisciplinary collaboration across robotics, AI, aerospace, and applied domain sciences.

\section{Foundations of Agentic UAVs}

\subsection{Architectural Components}

\begin{figure}[t]
     \centering
     \includegraphics[width=0.99\linewidth]{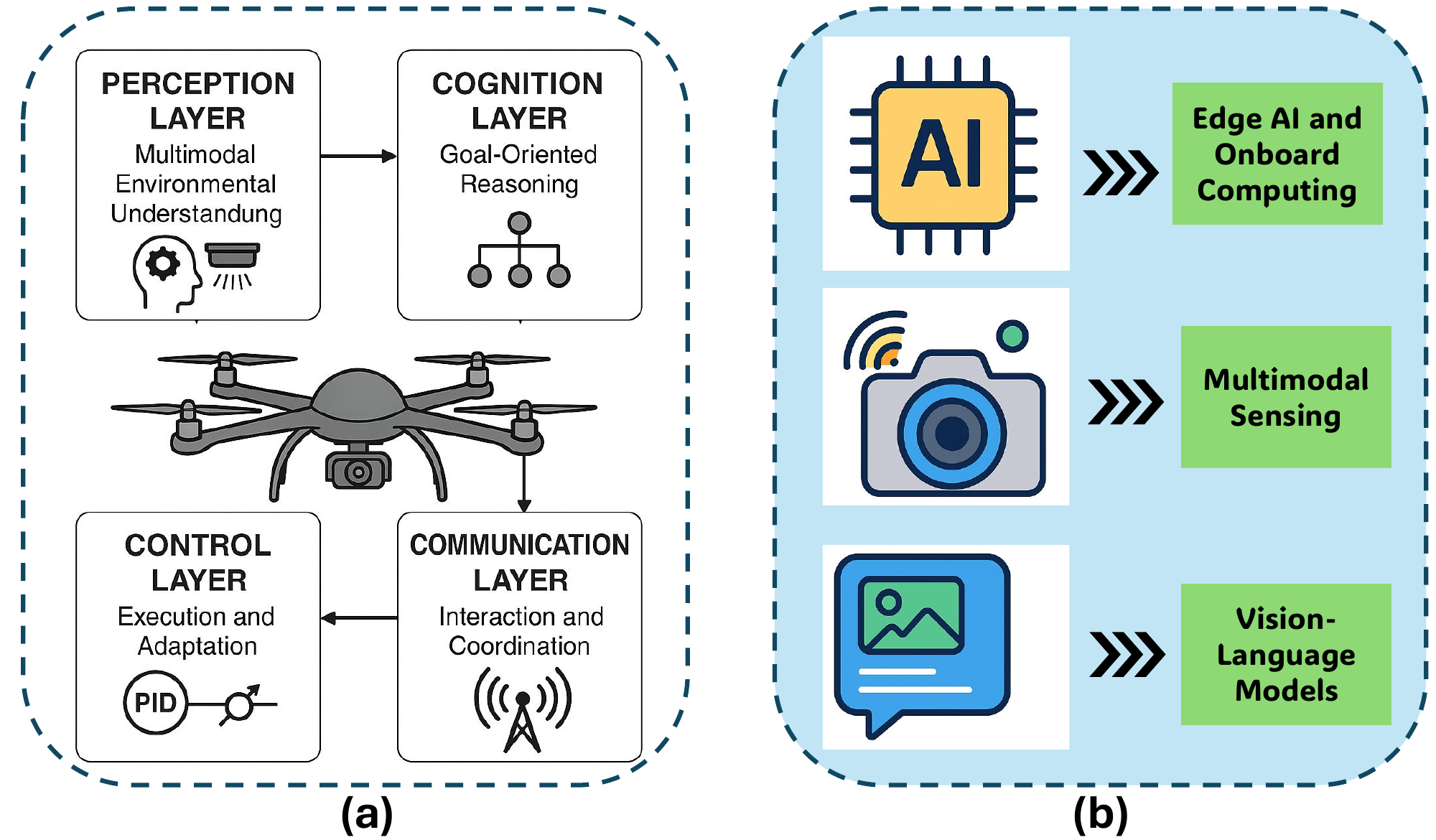}
    \caption{Illustration of the foundational elements of agentic UAV systems. (a) The architecture of agentic UAVs is structured into four key layers perception, cognition, control, and communication which collectively enable autonomous sensing, reasoning, acting, and interacting in dynamic agricultural environments. (b) Key enabling technologies that underpin these systems include edge AI and onboard computing, multimodal sensing for environmental perception, and    vision-language models (VLMs)  for instruction following and semantic decision-making, empowering UAVs to operate as intelligent aerial agents.}
    \label{fig:agentic_architecture_uav}
\end{figure}

The architecture of an agentic UAV (Figure \ref{fig:agentic_architecture_uav}a) is fundamentally organized around a hierarchical stack composed of four core layers: \textit{perception}, \textit{cognition}, \textit{control}, and \textit{communication}. Together, these layers enable the UAV to sense, reason, act, and interact autonomously in dynamic environments. This layered framework aligns with the vision-language-action (VLA) paradigm, wherein multimodal inputs are semantically interpreted and mapped to goal-directed outputs. Table \ref{tab:architectural_comparison} shows the architectural comparison between traditional and Agentic UAVs. 

\begin{table*}[htbp]
\centering
\caption{Architectural Comparison of Traditional vs Agentic UAVs}
\label{tab:architectural_comparison}
\begin{tabular}{|p{0.10\textwidth}|p{0.14\textwidth}|p{0.14\textwidth}|p{0.15\textwidth}|p{0.09\textwidth}|p{0.09\textwidth}|p{0.14\textwidth}|}
\hline
\textbf{UAV Type} & \textbf{Perception Modality} & \textbf{Control Architecture} & \textbf{Decision System} & \textbf{Autonomy Level} & \textbf{Task Adaptability} & \textbf{Communication Interface} \\
\hline

\textbf{Traditional UAVs} 
& Monocular or stereo RGB sensors; basic multispectral or thermal \cite{peng2020rgb, taha2019machine}; limited semantic parsing \cite{peng2020uav} 
& Rule-based flight controllers; manual or waypoint-following autopilots \cite{ozalp2013optimal, sampedro2016flexible} 
& Deterministic, scripted logic with no inference capabilities \cite{barnes2017humans, shin2016uav} 
& Level 1–2 (Basic autonomy; human-in-loop essential) \cite{yildirim2018system, nonami2010autonomous} 
& Static tasks and pre-defined operations with no reactive planning \cite{zhou2018mobile, xiaohuan2020aggregate} 
& Line-of-sight, unidirectional telemetry or ground control station (GCS) radio \cite{zeng2019accessing, rahmadhani2018lorawan} \\
\hline

\textbf{Agentic UAVs} 
& Multimodal sensing: RGB, thermal, LiDAR, hyperspectral; VLM-enabled semantic grounding \cite{tian2025uavs, feng2025multi, hallyburton2025trust} 
& Layered agentic control loops integrating perception, planning, memory, and reflection \cite{chitra2025artificial, sindiramutty2025swarm} 
& RL-based decision engines, memory-augmented modules, affordance-aware reasoning \cite{tang2025dnn, hao2025uav, Du2025} 
& Level 4–5 (Context-aware autonomy, minimal human supervision) \cite{wu2025context, anjum2024contextbots} 
& Real-time re-prioritization and dynamic adaptation during flight \cite{zhao2025flexifly, alqefari2025multi} 
& V2X networks, swarm-level coordination, edge-cloud sync \cite{Saad2025, Shahkar2025, sindiramutty2025swarm} \\
\hline

\end{tabular}
\end{table*}

\textbf{1. Perception Layer: Multimodal Environmental Understanding}

The perception layer is responsible for acquiring and preprocessing sensor data from the environment \cite{tian2025uavs}. A UAV’s sensor suite typically includes RGB and multispectral cameras, LiDAR, thermal sensors, inertial measurement units (IMUs), and barometers \cite{meng2023uav, balestrieri2021sensors}. Let $\mathbf{s}_t \in \mathbb{R}^n$ denote the sensor input vector at time $t$. The perception module maps raw data $\mathbf{s}_t$ to a semantic representation $\mathbf{o}_t$ via a function:
\[
\mathbf{o}_t = \Phi(\mathbf{s}_t)
\]
where $\Phi(\cdot)$ may be realized by convolutional neural networks (CNNs), transformers, or multimodal encoders trained for segmentation, detection, or affordance estimation. For instance, detecting chlorotic patches in crop fields or identifying survivors in a disaster zone are both outputs of $\Phi$.

\textbf{2. Cognition Layer: Goal-Oriented Reasoning}

The cognition layer transforms observations into decisions \cite{azuma2006review, wang2007cognitive, franklin2013lida, castelfranchi1998modelling} through reasoning, task decomposition, and planning \cite{hexmoor2025behaviour, dehghan2024review, chitra2025artificial}. Given a goal $g$ and observation $\mathbf{o}_t$, the cognitive policy $\pi$ computes the best action $a_t$:
\[
a_t = \pi(g, \mathbf{o}_t)
\]
This policy may be learned via reinforcement learning (RL) \cite{cui2019multi, ferdowsi2021neural, mondal2025multi}, where UAVs optimize cumulative reward $R = \sum_{t=0}^T \gamma^t r_t$, subject to environmental feedback. For example, in a seeding task, the agentic UAV must infer high-need planting zones from vegetation indices and plan a flight route to execute the task efficiently while minimizing energy consumption.

Planning modules in the cognition layer use spatial maps \cite{gupta2017cognitive},  $\mathcal{M}$ and goal priors \cite{abel2015goal} $P(g|\mathcal{M})$ to dynamically select trajectories or reconfigure tasks. Additionally, transformer-based architectures can incorporate temporal memory to support reflective control, where past missions inform current policy updates \cite{yu2025hybrid, mohseni2025robust}.

\textbf{3. Control Layer: Execution and Adaptation}

The control layer converts planned actions into executable trajectories. Given a UAV dynamics model $\dot{\mathbf{x}} = f(\mathbf{x}, \mathbf{u})$, where $\mathbf{x}$ is the state vector and $\mathbf{u}$ is the control input, the control system must generate $\mathbf{u}_t$ such that the UAV follows the trajectory implied by $a_t$:
\[
\mathbf{u}_t = \Gamma(a_t, \mathbf{x}_t)
\]
$\Gamma(\cdot)$ typically implements feedback control via PID, MPC (Model Predictive Control), or deep neural policy networks. For instance, trajectory tracking during obstacle-aware spraying in heterogeneous crop fields requires the UAV to continuously adapt to wind disturbances and terrain variations using real-time feedback.

\textbf{4. Communication Layer: Interaction and Coordination}

The communication layer facilitates data exchange and task coordination. Agentic UAVs utilize V2X (Vehicle-to-Everything) protocols to interface with other UAVs, ground vehicles, or cloud infrastructure. High-level coordination relies on shared semantic maps and symbolic goals. In multi-agent settings, a UAV broadcasts local observations $\mathbf{o}_t^i$ and actions $a_t^i$ to other UAVs to enable decentralized decision-making:
\[
\{a_t^i\} \sim \arg\max_{\{a\}} \sum_i R^i(\mathbf{o}_t^i, a_t^i, \mathbf{o}_t^{-i})
\]
where $\mathbf{o}_t^{-i}$ are observations from other agents. Swarm-based pasture mapping or synchronized spraying operations are typical use cases.

\textbf{Integration: Toward Agentic Autonomy}

In a fully agentic UAV, these four layers operate in tight feedback loops \cite{tian2025uavs, feng2025multi}. Perception provides context, cognition reasons over goals, control executes fine-grained actions, and communication enables distributed intelligence \cite{feng2025multi}. Embodied VLA models (e.g., Flamingo\cite{alayrac2022flamingo}, LLaVA \cite{liu2024improved}-Drone) increasingly allow UAVs to follow natural language instructions such as “map erosion zones near water bodies” and translate them into executable action pipelines \cite{zhao2025flexifly}. This modular yet integrated architecture enables UAVs to exhibit proactive behavior, adjust to real-world uncertainties, and collaborate across domains, making them viable as intelligent aerial agents for remote sensing, intervention, and environmental decision-making \cite{chitra2025artificial, natarajan2025artificial, girgin2025edgeai}.

\subsection{Key Enabling Technologies}

The emergence of agentic UAVs is underpinned by a convergence of key enabling technologies as depicted in Figure \ref{fig:agentic_architecture_uav}b, that empower these aerial systems to operate autonomously, interpret complex environments, and make goal-directed decisions in real time. Four core technologies form the foundation of this transformation: edge AI, onboard computing, multimodal sensing, and vision-language models.

\textbf{1. Edge AI and Onboard Computing}

Traditional UAVs offload data for post-processing, limiting their ability to respond in real time \cite{luo2015uav, ntousis2025dual, han2021over, hwang2022decentralized, abushahma2019comparative, flak2021drone}. In contrast, agentic UAVs integrate edge AI modules compact neural inference engines deployed on embedded systems such as NVIDIA Jetson, Intel Movidius, or Apple Neural Engine \cite{tang2025dnn, hao2025uav}. These processors support deep learning inference on the fly, enabling UAVs to perform tasks such as semantic segmentation, object detection, and path reconfiguration in situ. The inference function $F_{\text{edge}}(\mathbf{s}_t)$ maps sensor input $\mathbf{s}_t$ to decisions $a_t$ without relying on a cloud connection:
\[
a_t = F_{\text{edge}}(\mathbf{s}_t) \quad \text{subject to } \tau_c < \delta
\]
where $\tau_c$ is the computation latency and $\delta$ is the real-time threshold (e.g., 100 ms). Efficient edge computation is critical for dynamic operations such as obstacle avoidance, anomaly tracking, and adaptive spraying.

\textbf{2. Multimodal Sensing}

Agentic UAVs achieve environmental understanding through multimodal sensor fusion \cite{panduru2025exploring, javaid2025explainable, tian2025uavs}. RGB imagery captures visible structure \cite{frey2018uav, senthilnath2016detection, weiss2017using, liu2024crop, pei2025segmenting}, multispectral sensors reveal vegetation stress \cite{zhang2019mapping, berni2009thermal, barbedo2019review, dash2018uav, kandylakis2020water}, LiDAR enables 3D reconstruction \cite{chiang2017development, li2024single, maboudi2023review}, and thermal imaging detects heat anomalies \cite{tanda2024infrared, bajic2023uav, pruthviraj2023solar, tanda2020uav}. Each sensor modality provides a different channel $C_i$, and their fusion $\mathcal{F}$ forms a joint observation tensor:
\[
\mathbf{o}_t = \mathcal{F}(C_1, C_2, ..., C_n)
\]
For example, in pasture management, thermal + RGB fusion enables livestock detection under partial occlusion \cite{alanezi2022livestock, dadallage2024mask}. Sensor fusion also increases robustness under varying illumination or weather conditions \cite{golcarenarenji2022illumination, zhang2022robust, yue2024multi}.

\textbf{3.    Vision-Language Models (VLMs) }

VLMs such as Flamingo \cite{alayrac2022flamingo}, LLaVA \cite{liu2023visual}, and OpenFlamingo \cite{awadalla2023openflamingo} empower UAVs with the ability to understand and execute natural language instructions. These models jointly encode image inputs $\mathbf{I}$ and language tokens $\mathbf{T}$ into a shared latent space $\mathbf{z} = f(\mathbf{I}, \mathbf{T})$ from which semantic actions are derived. For instance, a VLM-enabled UAV can respond to commands like “inspect solar panels for damage” or “find low-vegetation areas near the stream.”   Vision-Language Models (VLMs)  enable instruction following, zero-shot generalization, and interactive dialogue, significantly enhancing human-UAV collaboration \cite{javaid2024large, tian2025uavs, guo2025bedi}.

\subsection{Comparison with Traditional UAVs}

The transition from traditional UAVs to agentic UAVs marks a paradigm shift in aerial robotics from automated platforms executing predefined tasks to intelligent agents capable of autonomous reasoning, adaptive behavior, and dynamic mission planning \cite{chai2024cooperative}. This evolution has profound implications for how UAVs perceive environments, interact with users, and operate in complex real-world missions. At the core of this transformation lies a shift in autonomy levels, decision-making architecture, and system integration \cite{thangamani2024drones, sharma2025path}.

\begin{table*}[htbp]
\centering
\caption{Scientific Comparison of Traditional UAVs vs. Agentic UAVs (Functional and Operational Aspects)}
\label{tab:comparison}
\begin{tabular}{|p{4cm}|p{5cm}|p{5cm}|}
\hline
\textbf{Aspect} & \textbf{Traditional UAVs} & \textbf{Agentic UAVs} \\
\hline
\textbf{Autonomy Level} & Operator-assisted \cite{ozalp2013optimal, nonami2010autonomous}, pre-programmed routines \cite{yildirim2018system}  & Context-aware \cite{wu2025context, anjum2024contextbots}, reflective autonomy with minimal intervention \cite{chitra2025artificial} \\
\hline
\textbf{Mission Planning} & Static waypoints \cite{sampedro2016flexible, yang2015path}, GPS-bound paths \cite{causa2018multi, stecz2020uav} & Dynamic goal planning, mission-aware routing, and replanning \cite{sautenkov2025uav, zhao2025general} \\
\hline
\textbf{Perception System} & Single-modality sensing \cite{taha2019machine} (e.g., RGB \cite{peng2020rgb}) & Multimodal sensing fused with semantic reasoning \cite{tian2025uavs}  \\
\hline
\textbf{Decision-Making Paradigm} & Deterministic \cite{shin2016uav, ji2016robust}, operator-initiated \cite{barnes2017humans, donmez2010modeling} & Learned policies, environment-driven actions, affordance-aware \cite{tang2025dnn, chitra2025artificial, mondal2025multi}\\
\hline
\textbf{Onboard Intelligence} & Minimal ; offloaded to ground stations \cite{liu2020artificial} & Edge-AI with onboard processing for inference and feedback loops \cite{sindiramutty2025swarm} \\
\hline
\textbf{Sensor Fusion Strategy} & Sequential/manual sensor integration \cite{peng2020uav, oh2010multisensor} & Real-time, context-driven multimodal sensor fusion \cite{hallyburton2025trust, panduru2025exploring} \\
\hline
\textbf{Task Adaptability} & Predefined task flow \cite{xiaohuan2020aggregate, zhou2018mobile}; no deviation \cite{liu2020decentralized} & Mid-mission adjustment, re-prioritization, adaptive response \cite{zhao2025flexifly,tang2025dnn, alqefari2025multi} \\
\hline
\textbf{Energy and Coverage Efficiency} & Low; redundant or suboptimal paths \cite{liu2018energy, li2018energy} & Energy-aware path optimization with policy guidance \cite{archana2025energy, gasche2025energy, Hai2025} \\
\hline
\textbf{Communication Strategy} & Basic telemetry; often unidirectional \cite{zeng2019accessing, rahmadhani2018lorawan} & V2X communication \cite{Saad2025, Shahkar2025}, real-time peer and cloud coordination \cite{sindiramutty2025swarm} \\
\hline
\textbf{Human-UAV Interaction} & GUI or joystick \cite{de2019using, jie2017design}; command-line tasks & Multimodal interface: speech, vision, and semantic grounding \cite{feng2025multi, Jiang2025} \\
\hline
\textbf{Collaboration Capability} & Solo operation; no inter-agent behavior \cite{alotaibi2019lsar} & Cooperative swarm behavior, shared map and task logic \cite{Du2025, Alqudsi2025} \\
\hline
\textbf{Learning and Memory} & Stateless \cite{aggarwal2020path, zhang2019iot}, no cumulative learning \cite{sanjab2020game} & Continuous memory update \cite{hao2025uav} $M_t$, reinforcement and self-learning \cite{Du2025} \\
\hline
\textbf{Resilience and Recovery} & Reactive; requires manual recovery \cite{skulstad2015autonomous, erdos2013experimental} & Autonomous fallback, contingency plan activation \cite{Tang2025, hao2025uav} \\
\hline
\textbf{Instruction Execution} & Scripted \cite{real2020unmanned}; static parameters only \cite{akram2017security, shen2018calculation} & Language-conditioned and vision-grounded goal execution \cite{sautenkov2025uav, tang2025dnn} \\
\hline
\end{tabular}
\end{table*}

\textbf{Traditional UAVs: Reactive and Scripted}

Traditional UAVs as depicted in Table \ref{tab:comparison} typically operate under predefined mission plans that are programmed prior to flight using ground control software \cite{damilano2013ground, ramirez2018extending}. Navigation relies on GPS waypoints \cite{yang2021uav}, and tasks such as imaging, spraying, or mapping are executed based on static rules \cite{arafat2023vision, gyagenda2022review}. These UAVs exhibit little to no autonomy beyond low-level control and require continuous supervision for dynamic reconfiguration. Sensor data are often collected passively and analyzed post-flight, limiting real-time responsiveness \cite{padua2017uas, boroujeni2024comprehensive, sapkota2023towards}.

In such systems, the autonomy level is largely classified as “automated,” where decisions are rule-based and task-specific. For example, a traditional crop scouting drone may follow a fixed grid pattern and capture RGB imagery \cite{abbas2023drones, shahi2023recent}, leaving disease detection or health analytics to human analysts after the mission concludes. These platforms lack the situational awareness or flexibility required to adapt to unexpected conditions such as sudden weather changes, GPS signal loss, or moving obstacles \cite{khan2022emerging, mahmoudzadeh2024holistic}.

\textbf{Agentic UAVs: Intelligent, Context-Aware Aerial Agents}

Agentic UAVs extend beyond automation into autonomy and agency. These systems integrate cognitive AI models, multimodal sensing, and edge intelligence to enable real-time interpretation, reasoning, and decision-making \cite{javaid2025explainable, chen2024situation}. Unlike traditional UAVs, they operate based on environmental context and mission goals \cite{pascarella2013agent, sautenkov2025uav, sezgin2025scenario}, allowing them to reconfigure their tasks mid-flight, learn from outcomes, and coordinate with other agents in multi-UAV systems \cite{sai2023comprehensive, cetinsaya2024pid}.

In mission planning, agentic UAVs employ adaptive algorithms that respond to semantic cues and goal hierarchies \cite{ma2024mission, tang2025dnn}. They do not simply follow routes they plan actions dynamically based on sensed data, mission constraints, and learned heuristics \cite{kumar2025comprehensive, jonnalagadda2025efficient}. For instance, if an agentic UAV detects chlorosis in a crop zone during a health monitoring mission, it can deviate from its original path, perform targeted resampling, and update its intervention priority an ability fundamentally absent in traditional platforms \cite{ali2024ai, nahiyoon2024recent}.

The autonomy of agentic UAVs is often modeled using hierarchical policies, planning graphs, or reinforcement learning frameworks \cite{natarajan2025artificial}. Given a goal state $g$ and current environment state $s_t$, the policy $\pi(s_t, g)$ produces an optimal action $a_t$, updating over time based on internal memory $M_t$:
\[
a_t = \pi(s_t, g; M_t)
\]
This formalism supports dynamic mission planning under uncertainty, enabling real-time replanning, multi-agent collaboration, and energy-efficient behavior.

\textbf{Systemic Implications for Autonomous Aerial Intelligence}

The rise of agentic UAVs expands the scope of autonomous aerial intelligence from simple execution to higher-order cognition \cite{guo2025bedi, xu2025multi}. Table~\ref{tab:comparison} compares the architectural, computational, behavioral, and operational differences between traditional and agentic UAVs across 15 core dimensions. These distinctions underscore the enhanced autonomy, adaptability, and intelligence of agentic platforms.

\textbf{Case Illustration: Aerial Crop Stress Response}

\begin{figure}[t]
     \centering
     \includegraphics[width=0.69\linewidth]{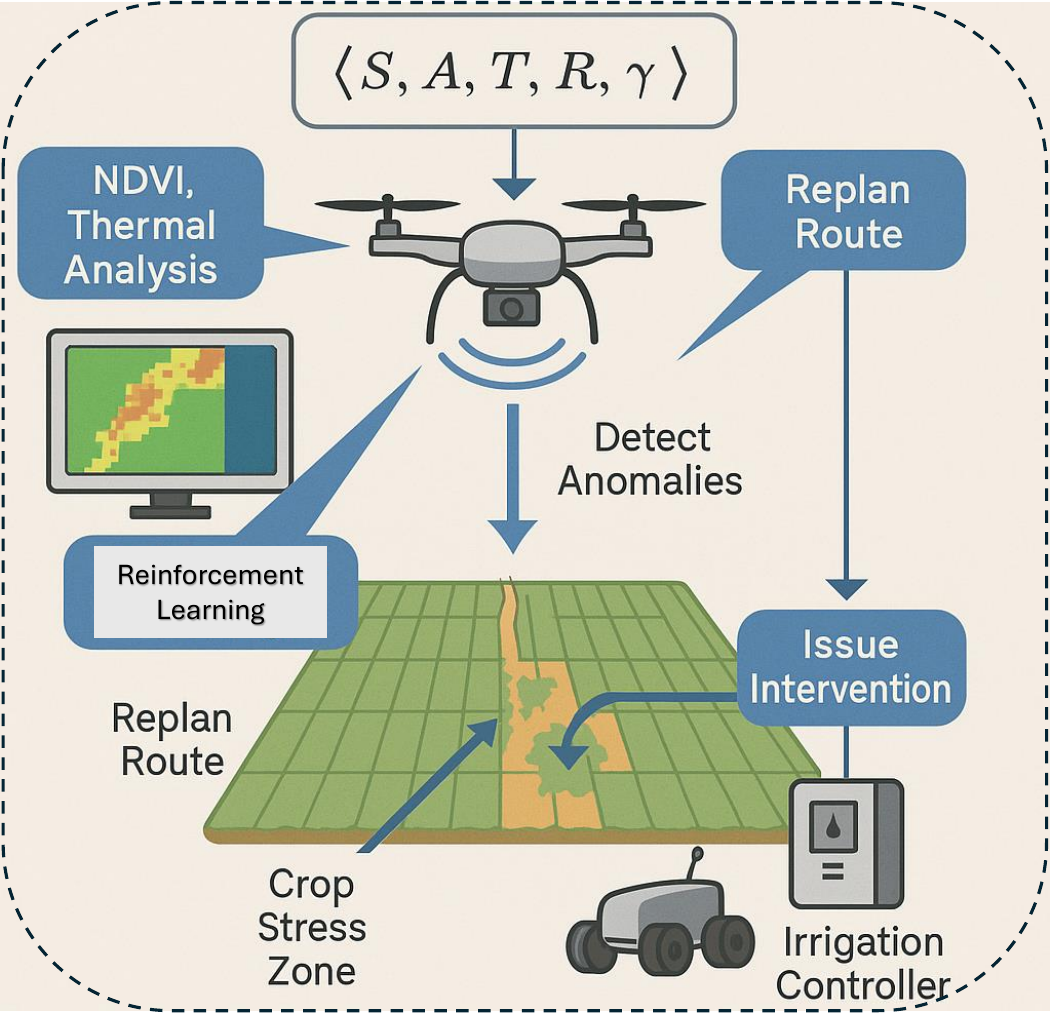}
    \caption{The diagram illustrates an agentic UAV autonomously monitoring a 100-hectare wheat field. Unlike traditional drones, it performs real-time NDVI and thermal analysis, detects chlorosis zones, adapts flight paths using onboard reinforcement learning, and issues irrigation commands. The architecture integrates multimodal sensors, semantic planning, onboard decision loops, and V2X communication with ground robots. Key elements include MDP-based reasoning, anomaly detection, energy-aware rerouting, and precision intervention highlighting how agentic UAVs enable intelligent, collaborative, and adaptive agriculture.}
    \label{fig:cropexample}
\end{figure}

Consider a UAV as illustrated in Figure \ref{fig:cropexample}, deployed in a 100-hectare wheat field for chlorosis detection. A traditional UAV would fly a fixed grid path, collect RGB images, and require post-flight human analysis. An agentic UAV, by contrast, would perform real-time NDVI and thermal analysis onboard, detect anomalies, replan its route to target stress zones, and issue intervention commands to an irrigation controller or ground robot.

Formally, its mission is modeled as a Markov Decision Process (MDP) \cite{puterman1990markov, garcia2013markov, white1989markov} defined by $\langle S, A, T, R, \gamma \rangle$, where:
- $S$: field state, crop health maps,
- $A$: flight actions (e.g., \textit{reroute}, \textit{rescan}),
- $T$: transition dynamics (e.g., wind disturbance),
- $R$: reward for accurate detection, minimal energy,
- $\gamma$: temporal discount factor.

Using reinforcement learning, the UAV adapts its action $a_t$ at time $t$ to maximize cumulative reward over mission duration:
\[
\pi^*(s_t) = \arg\max_{\pi} \mathbb{E} \left[ \sum_{t=0}^{T} \gamma^t R(s_t, a_t) \right]
\]
This intelligence loop illustrates the adaptive, proactive nature of agentic UAVs, elevating them from passive data collectors to real-time, collaborative aerial agents. The scientific and technological gap between traditional and agentic UAVs is more than architectural it reflects a cognitive divide. Agentic UAVs exhibit properties of autonomy, adaptability, and interpretability that define autonomous aerial intelligence. Their design prioritizes resilience to uncertainty, real-time decision-making, and alignment with mission semantics, making them indispensable in applications that demand speed, flexibility, and contextual awareness. As the field advances, the boundary between machine and intelligent aerial agent will continue to blur, ushering in a new era of scalable, agent-driven UAV systems across sectors.

\section{Multidomain Applications of Agentic UAVs}
\subsection{Precision Agriculture}

Agriculture has become one of the foremost domains to benefit from the integration of UAVs, driven by the need for scalable, sustainable, and data-informed farming practices \cite{velusamy2021unmanned}. Traditional agricultural UAVs have been widely used for aerial imaging \cite{olson2021review, gomez2014assessing, sapkota2021using}, pesticide application \cite{borikar2022application, quan2023economic}, and crop health monitoring \cite{ammad2018uav, devi2020review}. However, their reliance on predefined flight plans and post-flight analytics limits their responsiveness and adaptability in dynamic farm environments. Agentic UAVs address these limitations by introducing autonomy \cite{chitra2025artificial}, cognitive decision-making \cite{natarajan2025artificial}, and real-time mission adaptation \cite{tang2025dnn, sautenkov2025uav, sezgin2025scenario} thereby transforming the landscape of precision agriculture .

One of the most prominent applications is \textbf{crop health monitoring and mapping}. Agentic UAVs equipped with multispectral and thermal sensors, combined with onboard AI, can autonomously identify regions of chlorosis, pest outbreaks, or nutrient deficiency \cite{Crupi2025, Grando2025}. These UAVs dynamically adjust flight paths based on real-time vegetation indices such as NDVI and EVI, and they prioritize areas for closer inspection or revisit, using adaptive sampling. Unlike traditional UAVs, which capture imagery passively, agentic systems perform semantic analysis onboard and communicate insights to cloud-based farm management systems or ground-based actuators 

\textbf{Precision spraying} is another critical application, where agentic UAVs use AI-driven target identification and terrain awareness to apply agrochemicals selectively. These UAVs analyze plant health, wind conditions, and canopy structure in real time to optimize droplet size, application angle, and path efficiency \cite{Ye2025, Grando2025}. 

In the domain of \textbf{autonomous seeding and planting}, agentic UAVs operate in fragmented or topographically challenging fields where traditional machinery is infeasible. They use visual and topographic analysis to determine optimal seed release zones and dynamically adjust trajectories based on terrain slope, wind drift, and canopy occlusion. In regenerative agriculture and reforestation scenarios, these UAVs can sow cover crops or nitrogen-fixing species in underutilized plots with high spatial precision.

\textbf{Livestock and pasture monitoring} has also been transformed by agentic UAVs capable of detecting heat signatures, tracking movement patterns, and assessing animal health through thermal and multispectral analysis. Reinforcement learning-based policies allow UAVs to prioritize surveillance over stressed or isolated animals. In rotational grazing systems, agentic UAVs autonomously generate biomass maps and recommend paddock switching schedules based on forage conditions.

Finally, \textbf{environmental and resource monitoring} enables agentic UAVs to survey irrigation systems, map drainage performance, and detect emissions or soil degradation. These UAVs integrate microclimate data with multisensor inputs and autonomously prioritize inspections in high-risk zones, such as frost-prone areas or low-drainage fields. In connected farm ecosystems, they act as mobile observers that feed data into digital twin models of agricultural fields.

In summary, agentic UAVs as illustrated in Figure \ref{fig:cropexample} redefines precision agriculture by enabling proactive, context-aware, and mission-adaptive interventions across multiple farming tasks. Their integration into smart agriculture platforms promises higher productivity, input efficiency, and environmental sustainability.

\subsection{Disaster Response and Search-and-Rescue}

\begin{figure}[t]
     \centering
     \includegraphics[width=0.99\linewidth]{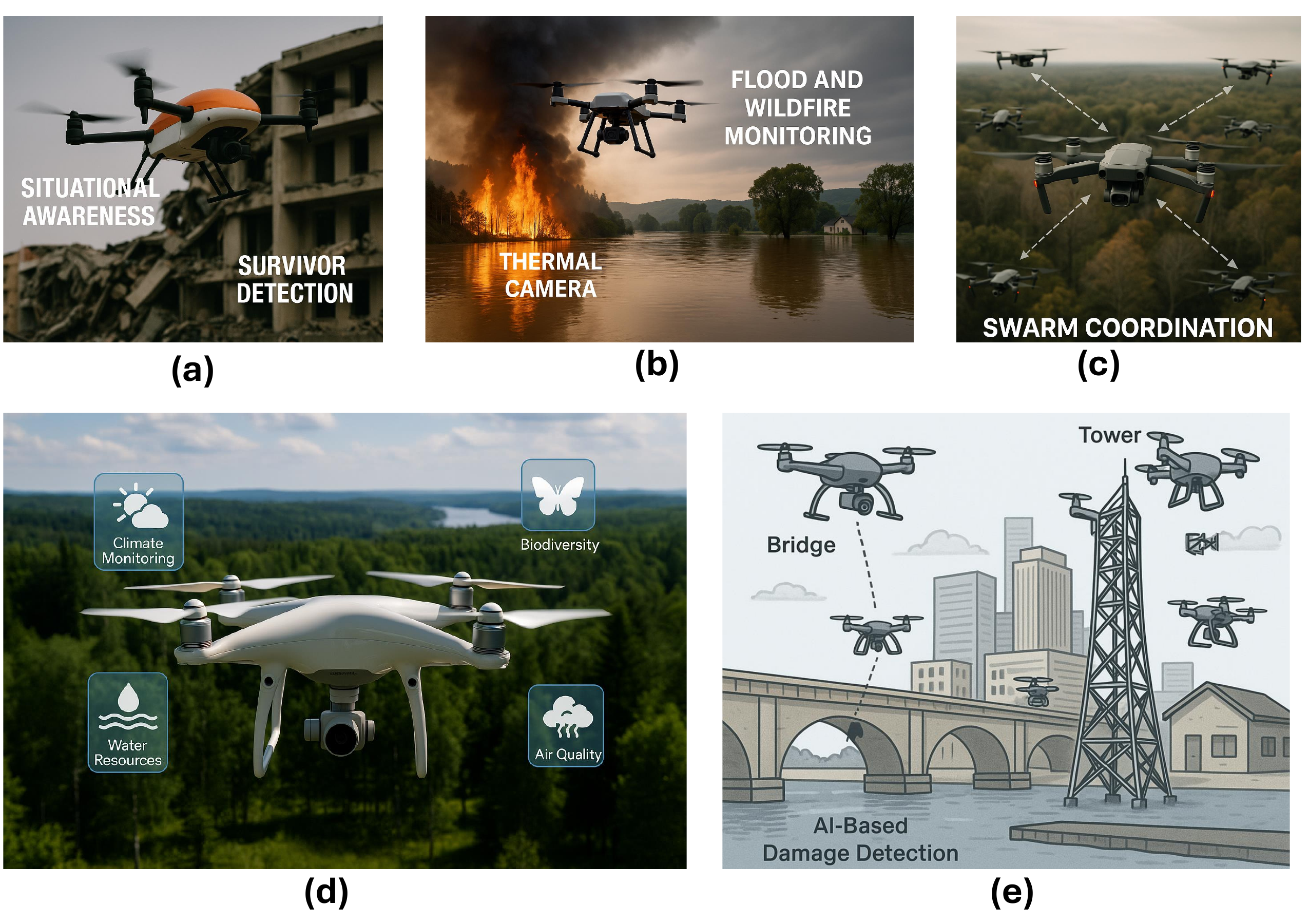}
    \caption{ Illustration showcasing multi-domain capabilities of agentic UAVs across real-world scenarios. (a) In disaster zones, UAVs autonomously perform situational awareness and survivor detection using multimodal sensing. (b) They adaptively monitor flood and wildfire conditions. (c) Swarm coordination enables efficient coverage through decentralized planning for environmental monitoring. (d) Environmental monitoring includes air, water, climate, and biodiversity sensing. (e) For urban infrastructure, UAVs conduct autonomous inspections of bridges and buildings using AI-driven anomaly detection. Together, these illustrate agentic aerial intelligence across critical sectors.}
    \label{appliocation1}
\end{figure}

Disaster response and search-and-rescue (SAR) operations represent one of the most critical domains for the deployment of agentic UAVs (Figure \ref{appliocation1}b). In these high-stakes, time-sensitive scenarios, traditional UAVs are often limited by static flight paths \cite{Alzahrani2020, Elmokadem2021, MahmoudZadeh2024}, operator dependence, and the lack of contextual reasoning. By contrast, agentic UAVs offer real-time adaptability, autonomous mission reconfiguration, and goal-driven behavior, making them uniquely suited for operations in dynamic, unstructured environments such as collapsed buildings, wildfire zones, or flood-affected regions \cite{Hickling2025, Gao2025}.

A foundational capability of agentic UAVs in this domain is \textbf{situational awareness}. Equipped with multimodal sensors such as RGB, thermal, and LiDAR agentic UAVs autonomously assess disaster zones, generate 3D terrain maps, and identify areas of structural instability. Using onboard SLAM (Simultaneous Localization and Mapping) and edge AI processors, these UAVs build live environmental models and update them continuously as new information is gathered \cite{Hickling2025}. In post-earthquake scenarios, for instance, UAVs can detect changes in elevation, debris fields, or building collapse patterns without relying on prior maps or GPS, which are often unavailable.

\textbf{Survivor detection} is another high-priority task where agentic UAVs excel. Through thermal imaging, movement tracking, and sound localization, these UAVs can identify potential survivors under rubble or in inaccessible terrain. Reinforcement learning algorithms help UAVs prioritize regions based on probability heatmaps of human presence, derived from patterns such as heat signatures near known shelter structures or movement patterns at the edges of debris fields.  VLMs  further allow operators to issue queries like “search for human shapes near the south quadrant” or “scan rooftops for waving hands,” which the UAV interprets and executes autonomously \cite{Hickling2025}.

In \textbf{flood and wildfire monitoring}, agentic UAVs provide dynamic mapping and hazard assessment. In wildfire zones, UAVs equipped with multispectral and thermal cameras can detect fire edges, embers, and high-risk ignition zones. Their ability to replan paths in real time based on wind direction or flame spread enables adaptive surveillance and firefighting coordination. In flood response, UAVs can assess water levels, map blocked evacuation routes, and identify stranded individuals using image segmentation and depth inference models.

\textbf{Swarm coordination} plays a pivotal role in scaling operations across large disaster zones. Agentic UAVs in a swarm operate under decentralized control, sharing information about coverage gaps, survivor sightings, and navigation hazards. Through vehicle-to-vehicle (V2V) communication and consensus-based planning, the swarm maintains collective awareness and adapts dynamically as individual agents encounter obstacles or complete sub-missions. This enables efficient area coverage, collision avoidance, and redundancy in case of hardware failures.

Finally, agentic UAVs contribute to \textbf{interoperable multi-agent response systems}. They coordinate with ground robots, emergency responders, and cloud platforms via semantic communication protocols \cite{Mequanenit2025, Mequanenit2025, Karim2025}. High-level mission goals such as “locate survivors in sector 4 and deliver water supplies” are decomposed into executable UAV behaviors, facilitated by cognitive AI and multimodal reasoning.

In summary, agentic UAVs significantly advance the speed, safety, and effectiveness of disaster response and SAR missions. Their autonomy, sensing intelligence, and collaborative behavior provide unparalleled advantages in scenarios where every second counts.

\subsection{Environmental Monitoring}

Environmental monitoring has become a global priority in the face of accelerating climate change, biodiversity loss, water scarcity, and increasing pollution. Traditional environmental sensing methods relying on stationary instruments, manual sampling, or satellite remote sensing often lack the spatial granularity, temporal frequency, and adaptive responsiveness needed to monitor rapidly changing ecosystems \cite{Manfreda2018, Panda2016, Mehedi2024}. Agentic UAVs, with their autonomous sensing, adaptive routing, and onboard AI capabilities \cite{natarajan2025artificial, tang2025dnn}, offer a transformative solution by acting as intelligent, mobile environmental sentinels.

A primary application of agentic UAVs in this domain is \textbf{climate-related ecosystem monitoring}  (Figure \ref{appliocation1}c and  (Figure \ref{appliocation1}d)). These UAVs autonomously measure microclimatic variables such as temperature, humidity, wind patterns, and solar radiation at high spatial and temporal resolution. In forest and agricultural landscapes, agentic UAVs use real-time weather sensors in combination with terrain-following control to map frost-prone zones, evapotranspiration patterns, or drought stress areas. Through reinforcement learning, UAVs can optimize flight trajectories to prioritize areas of rapid change or ecological risk \cite{Almalki2024}.

In \textbf{biodiversity conservation}, agentic UAVs are increasingly used for species detection, behavior observation, and habitat assessment. Thermal and hyperspectral sensors enable detection of wildlife under dense canopy cover, while acoustic sensors can be used to identify species-specific vocalizations. For instance, agentic UAVs have been deployed to monitor elephant and rhino populations in African reserves, using onboard object detection and thermal tracking to count individuals, assess movement patterns, and detect potential poaching risks. These UAVs adaptively focus on areas of high ecological value such as water holes or migration corridors and share data with conservation platforms in real time.

\textbf{Air quality surveillance} is another vital application. Agentic UAVs can be equipped with lightweight chemical sensors for detecting pollutants such as CO\textsubscript{2}, NO\textsubscript{x}, CH\textsubscript{4}, NH\textsubscript{3}, and particulate matter (PM2.5 and PM10). These UAVs autonomously navigate through industrial zones, agricultural fields, or urban neighborhoods, performing 3D plume mapping of emissions and correlating air quality with environmental or operational parameters. For example, in livestock production systems, UAVs can detect ammonia spikes over manure lagoons and trigger mitigation alerts. Swarm-based deployments can perform synchronized atmospheric sampling to characterize pollutant dispersion in complex terrains.

\textbf{Water resource monitoring} also benefits from agentic UAVs. Through spectral analysis and thermal imaging, these UAVs assess surface water quality, detect algal blooms, and monitor irrigation infrastructure \cite{Hong2025, Singh2025}. For example, UAVs operating over reservoirs and wetlands can detect changes in turbidity \cite{Cui2022, Trinh2024}, temperature gradients \cite{Elfarkh2023}, or surface scum indicative of eutrophication \cite{Pan2023, Rowan2021}. In precision irrigation systems, UAVs identify clogged emitters or uneven distribution zones and report anomalies to smart controllers for corrective action \cite{Mabrek2024, Suresh2025}. In flooded regions, agentic UAVs use depth inference and 3D reconstruction to map inundation extent and assist in risk assessment.

\textbf{Forest surveillance and carbon monitoring} represent long-term strategic applications. Agentic UAVs perform LiDAR-based 3D mapping to quantify biomass, canopy height, and tree density \cite{Chang2025, Li2025}. These measurements are essential for modeling carbon sequestration and tracking deforestation or degradation. Vision-language models enable natural language commands such as “map deforested patches near river boundary” or “monitor sapling growth in reforested zones,” allowing environmental researchers to deploy UAVs with minimal technical input. Over time, UAV-collected data feed into geospatial analytics platforms and ecological digital twins that inform climate action plans.

A distinctive strength of agentic UAVs in environmental monitoring is their ability to engage in \textbf{predictive, proactive behavior}. For instance, based on weather forecasts and satellite data inputs, a UAV may autonomously plan missions to assess pre-flood vulnerability zones, perform pre-fire vegetation dryness scans, or schedule follow-up flights in areas with past anomalies. Integration with Internet of Things (IoT) sensor networks allows UAVs to respond to alerts from ground-based sensors such as sudden temperature drops, river overflows, or gas leaks triggering targeted inspections with minimal latency.

Furthermore, these UAVs promote \textbf{participatory environmental sensing}. Through simplified interfaces and cooperative usage models \cite{AlRidhawi2021, Fei2022}, citizen scientists, farmers, or community groups can deploy UAVs for local environmental assessments. For example, community-led forest patrols in the Amazon now use UAVs to monitor illegal logging in protected areas, reducing human exposure to conflict zones and expanding the reach of conservation surveillance.

In summary, agentic UAVs offer an unparalleled combination of mobility, intelligence, and adaptability for environmental monitoring. By autonomously sensing, interpreting, and responding to ecological changes, these aerial systems not only improve data resolution and reaction time but also empower stakeholders at multiple scales from grassroots conservationists to global climate policy planners \cite{javaid2025explainable, Abro2024}.

\subsection{Urban Infrastructure Inspection}

The inspection and maintenance of urban infrastructure such as bridges, high-rise buildings, towers, roadways, and pipelines are critical to public safety, economic continuity, and disaster prevention. Traditional inspection methods often involve manual labor, scaffolding, or rope access, making them time-consuming, costly, and hazardous \cite{Mandirola2022}. While conventional UAVs have introduced some automation in data collection, they typically require GPS-based waypoint planning and operator control \cite{Tullu2021, Lawn2021, Morando2022}, limiting their effectiveness in complex or dynamic environments. Agentic UAVs, by contrast, provide a transformative approach by enabling autonomous, AI-driven inspection workflows with real-time damage detection and adaptive mission planning \cite{sezgin2025scenario}.

A core application of agentic UAVs in this domain is \textbf{bridge inspection}  (Figure \ref{appliocation1}e). Bridges require regular structural assessments for corrosion, crack propagation, and material fatigue especially in aging transport systems. Agentic UAVs equipped with high-resolution visual, infrared, and ultrasonic sensors can autonomously navigate complex bridge geometries, detect microcracks or spalling via computer vision models, and generate semantic 3D maps of damaged regions. Using simultaneous localization and mapping (SLAM) and deep learning-based defect classification, the UAV identifies key anomalies, localizes them within a digital twin model, and updates a maintenance schedule without requiring human supervision.

In the case of \textbf{high-rise buildings and towers}, agentic UAVs use terrain-aware planning and real-time path correction to conduct façade inspections. For instance, in urban centers such as New York or Singapore, these UAVs are deployed to autonomously scan skyscraper exteriors for facade cracks, sealant degradation, or broken windows. Multi-angle imaging combined with transformer-based vision models allows the UAV to detect defects that would otherwise be missed from limited perspectives. In telecom towers, UAVs autonomously inspect antenna positioning, cable integrity, and corrosion points, sharing real-time data with centralized asset management systems via cloud sync.

\textbf{Tunnel and under-bridge inspection} poses unique challenges due to GPS denial, confined spaces, and poor lighting. Agentic UAVs use LiDAR and visual-inertial odometry to maintain stable flight and map internal environments. They autonomously detect structural displacements, water seepage, and surface deformation using onboard segmentation models trained on tunnel-specific datasets. In metros like London and Tokyo, such UAVs are being piloted for nighttime infrastructure audits without interrupting daytime traffic or requiring track closures.

\textbf{Road surface monitoring} and \textbf{rooftop inspection} are also enhanced by agentic UAVs. For large-scale highways and urban pavements, UAVs autonomously detect potholes, rutting, and surface cracks using aerial CNNs and edge AI accelerators. In residential zones, UAVs perform autonomous inspection of solar panel alignment, roof shingle damage, and heat loss patterns from HVAC systems, using thermal cameras and spectral anomaly detection.

A unique strength of agentic UAVs is their ability to generate \textbf{automated inspection reports} and \textbf{anomaly heatmaps}. After data collection, onboard AI models segment images into defect classes (e.g., corrosion, misalignment, biological growth), quantify severity, and localize findings with centimeter precision. These reports are structured and compatible with civil engineering standards, allowing rapid review and decision-making.

In summary, agentic UAVs redefine urban infrastructure inspection by enabling autonomous, intelligent, and efficient evaluations of critical structures. Their ability to adapt to structural complexity, detect hidden defects, and operate without GPS or human oversight makes them indispensable tools for safe, smart, and scalable urban maintenance.

\subsection{Logistics and Smart Delivery}

The logistics and delivery sector is undergoing a rapid transformation fueled by advances in aerial autonomy, on-demand services, and urban mobility \cite{Kuru2019, BettiSorbelli2024, Jahani2025}. Traditional delivery UAVs have shown promise for last-mile transportation, particularly in rural areas or during emergencies \cite{Garg2023}. However, they are typically restricted by rigid path planning \cite{Jones2023}, reliance on GPS \cite{Grayson2018, Khan2021}, and the need for continuous human oversight \cite{Verdiesen2021}. Agentic UAVs, with their adaptive navigation, semantic understanding, and collaborative planning capabilities, are now emerging as intelligent aerial couriers capable of executing complex delivery tasks with minimal supervision \cite{dehghan2024review}.

\begin{figure}[t]
     \centering
     \includegraphics[width=0.99\linewidth]{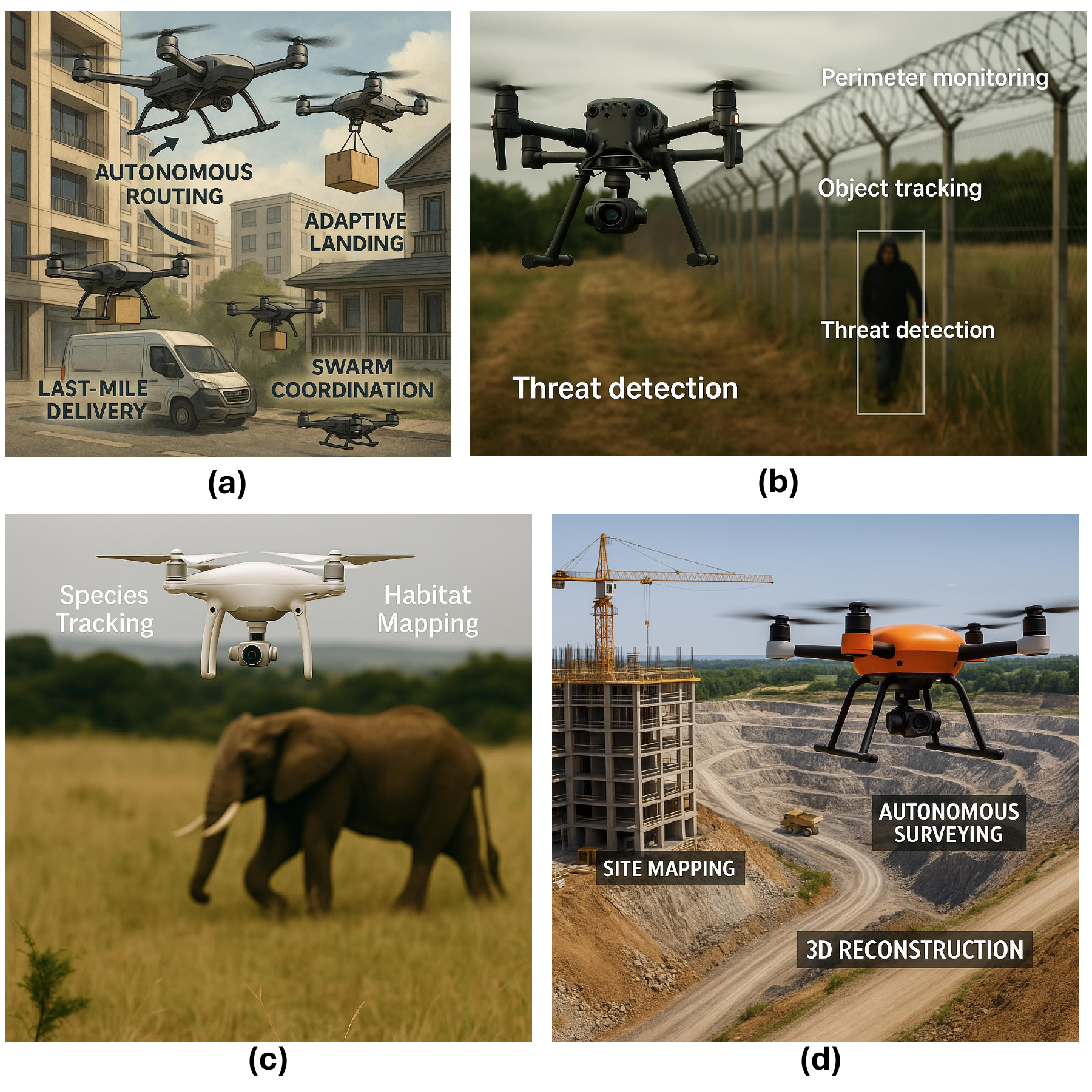}
    \caption{This illustration presents agentic UAVs operating across diverse real-world applications. (a) In logistics, UAVs perform autonomous last-mile delivery through adaptive routing and semantic landing. (b) For environmental and security monitoring, they conduct air quality sensing and autonomous perimeter surveillance. (c) In wildlife conservation, UAVs track species and detect poaching threats using thermal and visual analytics. (d) In construction and mining, they execute 3D site mapping, geospatial surveying, and infrastructure anomaly detection, enabling intelligent automation in industrial environments. }
    \label{appliocation2}
\end{figure}

A primary application is in \textbf{last-mile delivery} of medical supplies, e-commerce parcels, and critical documents as depicted in  (Figure \ref{appliocation2}a). In densely populated urban environments, agentic UAVs autonomously navigate through complex 3D spaces, avoiding buildings, power lines, and dynamic obstacles. These UAVs use multimodal perception fusing visual, LiDAR, and inertial data to perform GPS-denied navigation, even in narrow corridors between buildings. For example, during the COVID-19 pandemic, agentic UAVs were deployed in Rwanda and India to autonomously deliver vaccines and testing kits to remote clinics, with onboard AI rerouting them around weather systems or no-fly zones.

\textbf{Autonomous routing and dynamic re-tasking} are central features of agentic UAV logistics. Given a delivery goal $g$ and environmental state $s_t$, the UAV computes an optimal action $a_t = \pi(s_t, g)$ using reinforcement learning or behavior-tree policies. If the target location becomes unreachable due to weather, airspace conflict, or pedestrian congestion, the UAV dynamically replans using hierarchical mission graphs. Unlike traditional drones with fixed flight paths, agentic UAVs reason over delivery goals and prioritize based on time sensitivity, battery constraints, or package fragility.

\textbf{Adaptive landing and package drop-off} are handled using semantic scene understanding. Agentic UAVs detect suitable landing zones or drop sites such as open porches, balconies, rooftops, or designated delivery pads using instance segmentation and depth estimation. VLMs (vision-language models) further allow them to interpret instructions like “deliver to the yellow box near the entrance” or “leave package on rooftop helipad.” In low-light or crowded areas, the UAV adapts by requesting user confirmation via mobile alerts or QR code localization.

\textbf{Swarm coordination} is an emerging frontier in smart delivery, enabling multiple UAVs to cooperatively manage high-volume delivery tasks. In these scenarios, agentic UAVs share status, location, and payload information via V2V (vehicle-to-vehicle) communication and operate under decentralized coordination protocols. This allows swarms to reallocate tasks in real time, avoid airspace collisions, and minimize total energy expenditure. In pilot programs by Amazon Prime Air and Zipline, UAV fleets have demonstrated coordinated package delivery to multiple addresses in a single flight window using predictive delivery schedules and load balancing.

\textbf{Warehouse-to-doorstep integration} is also enhanced by agentic UAVs. These systems interface with autonomous ground robots and inventory management platforms, receiving package retrieval requests and synchronizing takeoff with real-time order queues. Using edge-cloud APIs, they dynamically optimize delivery sequences based on package volume, customer availability, and traffic conditions.

In summary, agentic UAVs extend aerial logistics beyond mere flight automation into the realm of intelligent, adaptive, and scalable last-mile delivery. Their ability to navigate uncertain environments, coordinate in fleets, and make real-time decisions positions them as essential components of next-generation logistics and urban mobility systems.

\subsection{Defense and Security Surveillance}

The defense and security sector demands high levels of reliability, autonomy, and situational awareness in challenging, high-risk environments \cite{Lyu2022, Hadi2023}. Traditional surveillance UAVs have long been used for reconnaissance \cite{Agbeyangi2016, Zhang2020, Rasmussen2017}, perimeter patrol \cite{Kalyanam2012, Krishnamoorthy2011}, and visual monitoring \cite{Ham2016}; however, their reliance on human operators, static flight plans, and limited onboard reasoning restricts their ability to respond to rapidly evolving threats. Agentic UAVs introduce a new paradigm in military and civilian security by enabling autonomous threat detection, collaborative patrolling \cite{Yang2024, tian2025uavs}, and semantic reasoning \cite{Lin2025, Liu2025}, thereby transforming aerial surveillance into an intelligent, mission-adaptive capability.

A key application is \textbf{autonomous perimeter monitoring} of military bases (Figure \ref{appliocation2}b), critical infrastructure, national borders, and restricted zones. Agentic UAVs equipped with thermal, infrared, and night vision sensors perform real-time surveillance, autonomously scanning for unauthorized entries, abnormal heat signatures, or motion patterns. Unlike conventional drones that follow static patrol paths, agentic UAVs dynamically replan their routes based on environmental cues or sensor feedback. For example, if an unidentified heat source is detected near a restricted fence line, the UAV autonomously adjusts its trajectory, zooms in with optical cameras, and begins high-frequency re-scanning for confirmation.

\textbf{Object tracking and classification} are also enhanced through onboard computer vision models trained for military targets such as vehicles, personnel, or weapons \cite{cazzato2020survey, liu2022military, laghari2024unmanned, criollo2024classification}. Agentic UAVs use deep neural networks and probabilistic filters to track moving objects under occlusion, varying illumination, or terrain interference \cite{din2025adversarial, kalafatidis2025swarmcatcher, natarajan2025artificial}. For instance, in forward operating zones, UAVs can follow suspicious vehicles or individuals across fragmented landscapes, handing off tracking responsibilities to other UAVs or ground assets when line-of-sight is lost. These systems support multi-target tracking and trajectory prediction, reducing the need for continuous remote piloting.

In \textbf{threat detection and anomaly assessment}, agentic UAVs apply reinforcement learning and anomaly segmentation techniques to identify deviations from expected patterns. This includes detecting unusual movement, object appearance, or sound patterns in sensitive zones. In counter-terrorism scenarios, UAVs have been used to identify unattended bags, loitering individuals, or unauthorized gatherings by comparing live inputs with behavior priors stored in semantic maps. When anomalies are detected, the UAV initiates an alert protocol, zooms in for validation, and communicates findings to command centers via secure channels.

\textbf{Autonomous patrols} are a significant advancement enabled by agentic UAVs. Patrol UAVs operate continuously or at programmed intervals with complete autonomy, scanning large territories without operator input. In border surveillance, for instance, agentic UAVs fly terrain-aware paths that adapt to environmental changes such as fog, wind, or topographical occlusion. Swarms of UAVs can autonomously divide coverage areas and perform persistent surveillance using decentralized consensus protocols. This capability is currently under trial in the EU and U.S. southern borders for 24/7 autonomous border security.

Moreover, \textbf{multi-agent coordination} plays a critical role in security operations. Agentic UAVs operate as distributed aerial nodes that synchronize with ground-based vehicles, stationary sensors, and human operators. In coordinated missions, one UAV may perform wide-area scanning, while others conduct close-in inspection, relay signals, or act as decoys. Vehicle-to-everything (V2X) communication ensures real-time updates and coordination, even under low-bandwidth or GPS-denied conditions. In complex operations such as hostage rescue or chemical spill containment, UAVs autonomously prioritize zones, monitor escape routes, and guide personnel through safe paths.

Another emerging application is in \textbf{infrastructure and airspace security}. Agentic UAVs are being deployed at airports, power plants, and ports to prevent sabotage or detect rogue drone intrusions. These systems perform high-frequency scans of airspace and infrastructure perimeters, using radar and RF triangulation to identify unauthorized aerial objects. In case of detection, agentic UAVs can autonomously intercept, escort, or disable intruding UAVs using kinetic or non-lethal mechanisms.

Agentic UAVs also support \textbf{adaptive mission control and human-in-the-loop integration}. Through voice interfaces, tablet-based planning tools, or VLM-assisted command, security officers can

\subsection{Wildlife Conservation and Ecology}

Wildlife conservation and ecological monitoring present unique challenges due to the remoteness, complexity, and sensitivity of natural ecosystems \cite{willis2015remote}. Traditional field methods for species tracking, habitat mapping, and anti-poaching surveillance are labor-intensive, time-consuming, and often limited in spatial scale \cite{goyal2025limitations, santhosh2025integration}. While UAVs have provided substantial improvements in aerial surveying, their functionality has historically been restricted to passive data collection and operator-defined flight plans. Agentic UAVs overcome these limitations by enabling autonomous, context-aware, and ethical monitoring of ecosystems with minimal disturbance to wildlife behavior.

One of the most prominent applications of agentic UAVs is in \textbf{species tracking and behavioral observation} (Figure \ref{appliocation2}c). These UAVs are equipped with thermal imaging, hyperspectral cameras, and acoustic sensors that can detect and identify animals under dense canopy, at night, or in rugged terrain. Using reinforcement learning, UAVs prioritize patrols in biodiversity hotspots, known migration routes, or areas with ecological stress. For example, in the Serengeti and Amazon rainforest, agentic UAVs have been used to autonomously locate and follow elephants, big cats, and primates, learning to adjust altitude and flight behavior to minimize disturbance. Through real-time tracking, conservationists gain insights into population dynamics, foraging patterns, and seasonal migrations.

\textbf{Anti-poaching surveillance} is another critical function where agentic UAVs excel. Traditional UAVs require human operators to detect and interpret suspicious activity, which delays response time. In contrast, agentic UAVs use vision models trained to recognize poacher behavior such as armed individuals, fire usage, or unusual movement near protected zones and autonomously alert ground teams. In several African reserves, these UAVs operate at night, detecting poachers using thermal anomalies and relaying coordinates to ranger units within seconds. Their ability to coordinate in swarms also enables wide-area coverage and persistent aerial vigilance, reducing illegal hunting activities dramatically.

In \textbf{habitat and ecological mapping}, agentic UAVs autonomously collect high-resolution spatial data used to monitor changes in land cover, vegetation health, water availability, and fragmentation. These UAVs generate 3D habitat models, detect invasive plant species using multispectral imagery, and map nesting or breeding sites of endangered species. Through temporal analysis and semantic segmentation, they identify long-term trends such as deforestation, desertification, or coral reef bleaching. In wetlands and coastal zones, UAVs have been deployed to assess mangrove health and track migratory bird populations during critical breeding seasons.

A unique advantage of agentic UAVs is their capacity for \textbf{ethical drone-animal interaction}. Traditional drones often disrupt wildlife due to their noise or invasive flight patterns \cite{Afridi2025, Mo2022}. Agentic UAVs, however, use animal behavior modeling and proximity-aware flight control to minimize stress \cite{Sellami2025}. By monitoring body language cues such as agitation, vocalization, or group dispersion UAVs adjust their distance, altitude, or speed in real time to remain unobtrusive. This ethical design principle ensures that UAVs support non-invasive data collection and wildlife welfare, which is increasingly important in ecological research ethics.

Moreover, these UAVs promote \textbf{community-led conservation}. Simplified interfaces and natural language commands allow indigenous groups and local rangers to deploy UAVs for resource monitoring, boundary patrols, and biodiversity assessments. By automating data analysis and report generation, agentic UAVs reduce the technical barriers to ecological stewardship. Thus, agentic UAVs empower wildlife conservationists and ecologists with intelligent, adaptive, and ethical aerial systems. Their ability to autonomously detect species, prevent poaching, monitor habitats, and interact responsibly with animals marks a significant step forward in sustainable conservation technology.

\subsection{Construction and Mining Automation}

Construction and mining operations involve large-scale spatial planning, real-time resource management, and safety-critical workflows \cite{AlMarri2025, Obosu2025}. Traditional UAVs have played a valuable role in aerial imaging and volume estimation, but they often lack the cognitive flexibility required to operate autonomously in dynamic, high-risk industrial environments \cite{Mohsan2023}. Agentic UAVs, powered by onboard AI, real-time decision-making, and multimodal perception \cite{javaid2025explainable}, offer a transformative solution supporting autonomous surveying, site intelligence, and operational optimization across the lifecycle of construction and mining projects.

One of the primary applications is \textbf{site mapping and 3D reconstruction} as depicted in Figure \ref{appliocation2}d. Construction sites undergo frequent changes due to excavation, material delivery, and structural progress. Agentic UAVs autonomously monitor and document these changes using stereo vision, LiDAR, and photogrammetry. They generate real-time 3D point clouds and semantic models \cite{Chitta2024, Mehranfar2024},  that integrate into Building Information Modeling (BIM) systems. For example, in smart construction projects in Japan and the UAE, agentic UAVs fly daily missions to update site geometry, detect deviations from design blueprints, and assess excavation volumes, thereby enabling proactive adjustments to scheduling and material allocation.

In \textbf{mining operations}, particularly open-pit and aggregate mines, agentic UAVs are used for \textbf{autonomous surveying and geospatial analysis} \cite{Mitchell2017}. These UAVs compute optimal flight paths based on terrain topology and dynamically adjust their altitude and viewing angle to ensure consistent data capture. They measure bench heights, slope angles, and pit depth using LiDAR and RTK GPS, and detect geological hazards such as wall instability or overburden failure. In Australia’s iron ore mines, for instance, agentic UAVs continuously update digital terrain models and provide mining engineers with automated cut-and-fill volume reports, improving both safety and efficiency.

\textbf{Inventory and stockpile monitoring} is another critical use case. In cement plants, quarries, and construction yards, agentic UAVs autonomously quantify volumes of aggregates, sand, and raw materials using computer vision and volumetric analysis. These UAVs compare current stockpile volumes with historical baselines to detect overuse or understocking, issuing alerts to logistics managers. Unlike traditional UAVs requiring manual analysis, agentic systems generate labeled 3D reconstructions and automate reporting, integrating seamlessly into ERP or SCM software for real-time resource tracking.

\textbf{Progress tracking and quality inspection} are enhanced by the agentic UAV’s ability to semantically interpret visual scenes. UAVs recognize structural elements such as beams, columns, and walls, and assess their completion status by comparing current imagery with construction schedules. AI-based defect detection models identify misalignments, cracks, or missing components early in the construction cycle. In precast concrete and steel structure projects, agentic UAVs automatically detect alignment errors and flag them for human verification, reducing costly delays and rework.

\textbf{Autonomous safety inspections} are also supported, particularly in hazardous or confined environments such as tunnels, scaffolding, or heavy machinery zones. Agentic UAVs monitor worker presence, detect PPE (personal protective equipment) compliance, and identify hazardous conditions like material spills, loose wiring, or unstable structures \cite{Jordan2018, Li2023}. Reinforcement learning models allow UAVs to adapt their inspection strategies based on site risk maps and near-miss incident reports \cite{Wu2024, Pantaleon2025}. In deep mining operations, where ground-based inspection is dangerous, these UAVs navigate narrow shafts and stope entries to check for gas leaks or structural stress.

\textbf{Drone-to-machine integration} enhances automation by linking UAV outputs with ground-based machinery. Agentic UAVs communicate excavation depth maps to autonomous bulldozers or graders, dynamically updating task boundaries. In advanced construction robotics frameworks, UAVs direct robotic arms or cranes for optimal material placement based on real-time spatial awareness. In large-scale mining, agentic UAVs guide autonomous haul trucks to optimize load routes and avoid congestion.

Moreover, these UAVs support \textbf{long-term project analytics}. Using computer vision and metadata tagging, they record historical timelines of construction progress and provide visual documentation for regulatory compliance, dispute resolution, and stakeholder engagement. Natural language interfaces also allow project managers to issue high-level commands like “survey the northwest stockpile” or “check wall alignment on level 3 of the west structure,” which the UAV interprets and executes autonomously.

\section{Challenges and Limitations}
\subsection{Technical and Operational Constraints}

\begin{figure}[t]
     \centering
     \includegraphics[width=0.99\linewidth]{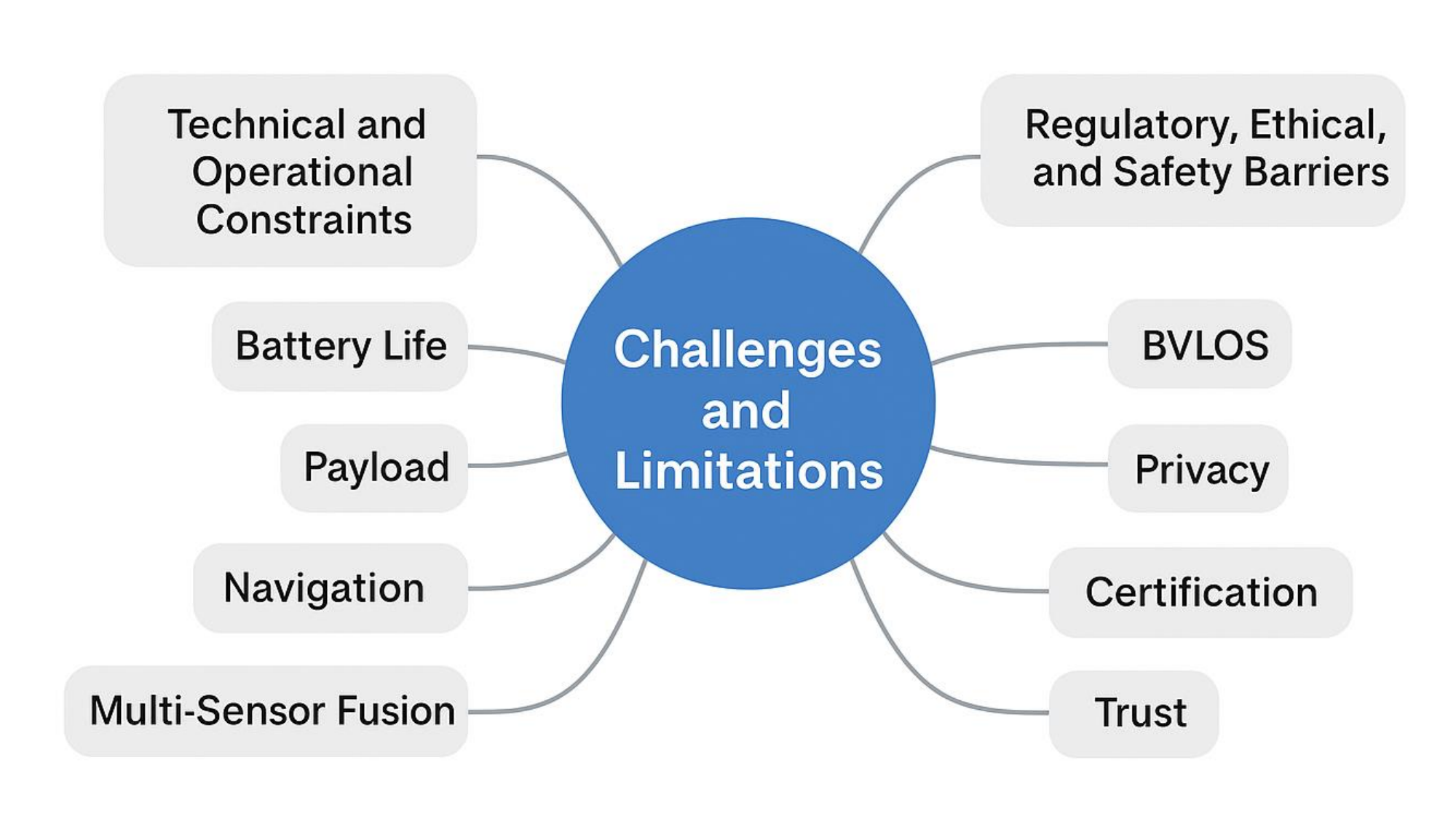}
    \caption{This scientific diagram illustrates ten key barriers to the adoption of agentic UAVs. On the technical side, issues include battery life, payload capacity, navigation, sensor fusion, and compute limitations, all of which constrain autonomous performance and real-time processing. On the regulatory and societal side, limitations include BVLOS (Beyond Visual Line of Sight) regulations, privacy, certification, trust, and model generalization, each posing challenges to safety, accountability, and scalable deployment in real-world environments.}
    \label{challengeLimit}
\end{figure}

Despite the transformative potential of Agentic UAVs across diverse applications, their widespread deployment is hindered by several technical and operational constraints  as depicted in Figure \ref{challengeLimit}. These challenges arise from the need to balance onboard intelligence with hardware limitations, mission-specific performance requirements, and environmental robustness \cite{sezgin2025scenario, Karbasishargh2024}. While advancements in autonomy and decision-making have improved the capability of UAVs to operate in complex scenarios, the physical and computational limitations of aerial platforms impose critical bottlenecks \cite{MahmoudZadeh2024}.

One of the most pressing limitations is \textbf{battery life and energy efficiency}. Most small- to medium-scale UAVs operate using lithium-polymer (LiPo) or lithium-ion batteries, offering limited flight durations ranging between 20 to 45 minutes under standard payload conditions \cite{Kang2023, SchachtRodriguez2018}. Agentic UAVs, which require additional energy for onboard AI inference, sensor fusion, and real-time communication, consume power more aggressively than conventional UAVs \cite{natarajan2025artificial, kourav2025artificial}. Edge computing workloads, such as object tracking or SLAM, significantly increase energy demands. This results in constrained mission time, particularly in applications requiring persistent surveillance, long-range delivery, or real-time environmental monitoring. While hybrid power solutions (e.g., solar-assisted or fuel-cell UAVs) are being investigated, they have yet to meet the energy density requirements for high-agility, AI-driven aerial operations \cite{CaballeroMartin2024}.

\textbf{Payload capacity and sensor integration} present another critical bottleneck \cite{krishnan2022roofline, mohsan2023unmanned, wan2024advancements}. Agentic UAVs require multiple sensors visual, thermal, LiDAR, radar, and acoustic to perceive and interpret their surroundings robustly. However, each additional sensor increases weight, power consumption, and thermal footprint, directly affecting flight time and maneuverability. Lightweight sensors often compromise on resolution or range, thereby reducing perceptual fidelity. Furthermore, integrating diverse sensors into a limited payload space without introducing electromagnetic interference or destabilizing aerodynamics is a non-trivial engineering challenge. This trade-off between sensing richness and flight efficiency restricts UAV deployment in tasks requiring complex multi-modal awareness, such as urban infrastructure inspection or underground mine exploration.

\textbf{Real-time navigation and localization} in GPS-denied or cluttered environments remains a significant hurdle \cite{dissanayaka2023review, ramirez2022unmanned, zhukov2024enhancing}. While visual-inertial odometry (VIO), LiDAR SLAM, and acoustic localization offer alternatives to GPS, they are susceptible to drift, occlusion, and environmental noise. In dynamic environments like forests, construction sites, or disaster zones, consistent localization requires not only robust mapping but also high-frequency recalibration, which stresses onboard processing units. Additionally, integrating external signals such as ground beacons or map priors is often infeasible in field deployments. Autonomous path planning under these uncertain localization regimes can lead to suboptimal decisions, collisions, or mission failure.

\textbf{Multi-sensor fusion and synchronization} is central to agentic behavior but remains a deeply challenging task. Agentic UAVs must combine information from various modalities (e.g., visual, thermal, depth, IMU) to infer semantic understanding of the scene. However, sensors operate at different temporal and spatial resolutions and exhibit nonlinear noise characteristics. Aligning these signals in real time requires calibration pipelines, time-stamping synchronization, and outlier rejection algorithms all of which increase system complexity. In fast-changing conditions (e.g., smoke in fire zones, moving vehicles in cities), even minor fusion inaccuracies can degrade decision quality, obstacle avoidance, or anomaly detection.

\textbf{Onboard compute limitations} are a significant constraint, especially for small UAVs operating at the edge \cite{mahmoudzadeh2024holistic}. While modern UAVs may incorporate embedded GPUs (e.g., NVIDIA Jetson series), they still struggle with the computational demands of transformer-based vision-language models, high-resolution 3D reconstruction, or large-scale reinforcement learning policies. This leads to trade-offs between model complexity, frame rate, and task latency. Offloading computations to cloud systems is often not viable in remote or latency-sensitive applications. Additionally, the thermal management of compact compute modules under continuous load presents another bottleneck, risking component degradation or emergency shutdowns mid-flight.

Thus, agentic UAVs face an array of technical and operational constraints that limit their deployment across real-world, high-demand scenarios. Addressing challenges related to energy efficiency, payload balancing, real-time localization, sensor fusion, and compute scalability is critical for transitioning from pilot deployments to fully autonomous aerial intelligence systems. Future work must focus on co-optimizing hardware-software stacks, energy-aware AI models, and ultra-light, high-performance sensing systems to overcome these barriers.

\subsection{Regulatory, Ethical, and Safety Barriers}

While Agentic UAVs promise unprecedented autonomy and decision-making capabilities across various domains, their large-scale deployment remains significantly constrained by regulatory, ethical, and safety concerns. These challenges span beyond technical feasibility and touch on broader issues of airspace governance, public trust, data responsibility, and compliance with global norms \cite{shavit2023practices, raza2025trism}. Unlike conventional UAVs that function under tight human oversight, agentic systems make independent decisions, thereby raising novel concerns around accountability, certification, and societal acceptance.

A foremost challenge lies in the evolving \textbf{regulatory frameworks}, particularly for \textbf{Beyond Visual Line of Sight (BVLOS)} operations \cite{van2023beyond}. Most civil aviation authorities including the FAA (U.S.), EASA (Europe), and ICAO (global) still impose strict limitations on BVLOS flights \cite{hartley2022bvlos, mctegg2022comparative, safie2025regulatory}, which are essential for many applications such as long-range delivery, search-and-rescue, and wildlife monitoring. BVLOS operations require robust detect-and-avoid (DAA) systems, real-time telemetry, and backup communication protocols. Although agentic UAVs can autonomously navigate and make adaptive decisions, current regulatory standards do not recognize these systems as reliably certifiable under existing rules. The lack of standard benchmarks for autonomy levels, explainability, and redundancy hinders the approval process for autonomous aerial systems.

\textbf{Privacy and data governance} present another major obstacle. Agentic UAVs often collect high-resolution imagery, thermal scans, or behavior-sensitive data during autonomous missions. In urban or civilian contexts such as infrastructure inspection or law enforcement these UAVs may inadvertently record personally identifiable information (PII) or violate spatial privacy boundaries. The autonomous nature of their operation complicates consent and accountability: who is responsible if a UAV unintentionally films private property or tracks individuals without authorization? Existing privacy laws (e.g., GDPR in Europe or CCPA in California) are not tailored to autonomous data collectors that adapt their behavior in real time. Designing responsible AI mechanisms that limit unnecessary data collection, enforce geofencing, and implement onboard anonymization protocols is critical to ethical deployment.

\textbf{Certification and compliance} procedures for agentic systems are also underdeveloped. Traditional UAV certification frameworks evaluate safety based on mechanical reliability, operator training, and communication link stability. However, agentic UAVs rely on probabilistic decision policies, dynamic task assignment, and machine-learned behavior models, which do not fit cleanly into conventional certification pipelines. For instance, how should regulators validate the safety of a reinforcement learning policy that evolves over time or a VLM-based controller interpreting language commands? Furthermore, the interpretability of these systems remains limited raising concerns about failure analysis, traceability, and post-incident accountability.

\textbf{Trust and societal acceptance} are equally critical. The deployment of UAVs with high-level autonomy in public spaces often triggers concerns regarding loss of control, surveillance, and unintended harm \cite{perz2024multidimensional}. Public distrust can stem from both visible intrusiveness such as low-flying drones near residential areas and invisible autonomy, where people are unsure how UAVs make decisions \cite{froomkin2015self, chakravarti2021strategies}. For successful integration into civil infrastructure, agentic UAVs must be explainable, predictable, and transparent. This includes providing users and bystanders with interpretable summaries of UAV intent, current mission goals, and safety protocols in case of emergency landing or communication failure.

To mitigate these risks, the concept of \textbf{Responsible AI for agentic UAVs} is gaining traction. This includes designing systems that are fair, auditable, and accountable. Embedding ethical reasoning into decision models for example, deferring high-risk decisions to human operators, or rejecting commands that violate predefined ethical boundaries can prevent misuse or catastrophic failure. Moreover, human-in-the-loop (HITL) and human-on-the-loop (HOTL) frameworks are being explored to allow oversight without compromising agentic autonomy.

In conclusion, regulatory, ethical, and safety barriers present formidable challenges to the widespread adoption of Agentic UAVs. Overcoming these issues will require not only advances in technology but also multidisciplinary collaboration across aviation regulators, ethicists, AI researchers, and public policy experts. Establishing harmonized standards, ethical AI frameworks, and certifiable autonomy benchmarks is essential to ensure that the deployment of Agentic UAVs is both legally sound and socially beneficial.

\subsection{Data and Model Reliability Issues}

The functionality of agentic UAVs hinges critically on the robustness of their onboard models and the quality of sensor data used to interpret and interact with complex environments. These systems rely on computer vision, perception algorithms, language-grounded reasoning, and real-time planning to execute autonomous missions. However, the reliability of these AI-driven operations is often compromised by limitations in data diversity, model generalization, semantic misinterpretation, and computational uncertainty. Such issues not only affect performance but may also pose serious safety and mission risk in dynamic, real-world settings.

One of the most pressing concerns is \textbf{model generalization} \cite{tian2025uavs}. Many of the models deployed on agentic UAVs ranging from object detectors and semantic segmenters to multimodal  VLMs  are trained on datasets collected in controlled or idealized settings. These models often fail to generalize when exposed to unseen weather conditions, diverse geographical terrains, or mission-specific edge cases. For instance, an object detector trained on urban traffic data may struggle to recognize construction vehicles in a quarry or emergency responders in disaster zones. Similarly, semantic scene understanding may misclassify shadows as obstacles or fail to detect transparent or reflective surfaces such as glass panels or water bodies, leading to erroneous decision-making.

\textbf{Real-time inference reliability} is another major bottleneck in UAVs \cite{muzammul2024enhancing,lai2023real}. Agentic UAVs are expected to process high-throughput sensor data visual, thermal, depth, and inertial in real time to perform tasks such as tracking, navigation, and hazard detection. However, under field conditions, these inference pipelines may lag, crash, or return partial results due to hardware constraints or input noise. Dropped frames, sensor occlusions, or hardware overheating can result in delayed or inaccurate decisions. This latency is particularly critical in high-speed operations such as infrastructure inspection near power lines or low-altitude search-and-rescue missions, where milliseconds matter for safe maneuvering.

\textbf{Semantic interpretation errors}, especially in VLM-enabled UAVs \cite{ye2025vlm, li2025navblip, cai2025flightgpt, yaqoot2025uav}, raise additional concerns. These models are designed to interpret language-based commands (e.g., “search near the south tree line” or “inspect the blue container on the rooftop”) and execute corresponding flight actions. However, these instructions may be ambiguous, context-dependent, or culturally specific, leading to incorrect execution. Furthermore, the VLM may lack sufficient grounding or world knowledge to understand nuances in spatial or task semantics. Misinterpretation of terms like “safe zone,” “boundary edge,” or “damaged structure” can lead to navigation into restricted areas, missed detections, or incomplete surveys.

\textbf{Uncertainty estimation and error quantification} are often missing or underdeveloped in current agentic UAV systems. Many AI models output deterministic decisions without expressing confidence levels or quantifying prediction uncertainty. In critical missions such as wildlife monitoring or disaster response, failure to convey uncertainty in species detection or terrain mapping may result in wasted resources or endanger human teams. While techniques such as Bayesian deep learning or Monte Carlo dropout can provide uncertainty estimates, they are computationally expensive and difficult to deploy on edge devices within size, weight, and power (SWaP) constraints.

Another issue is \textbf{training data reliability and annotation bias}. UAV perception models are often trained on datasets that lack comprehensive representation across domains (e.g., desert vs. forest, day vs. night, dry vs. rainy) \cite{li2021uav, wang2025diverse}. These datasets may also inherit labeling inconsistencies, sensor calibration errors, or class imbalance, leading to brittle performance in critical environments \cite{rahman2024uav}. The lack of large-scale, high-quality, domain-specific aerial datasets limits the robustness of general-purpose perception modules. Moreover, sim-to-real gaps persist between simulated training environments and the physical world, necessitating domain adaptation methods that are still in early research stages.

\textbf{Continual learning and model degradation} further complicate reliability. In long-duration deployments, agentic UAVs are exposed to drifting data distributions and evolving environments. Without continual learning or online adaptation mechanisms, model performance degrades over time. However, enabling safe and stable online learning in flight without risk of catastrophic forgetting or performance collapse is an open challenge, especially under constrained compute and memory budgets.

In summary, ensuring reliable data interpretation and model behavior remains one of the central challenges in deploying agentic UAVs at scale. Addressing issues of generalization, uncertainty, semantic grounding, and real-time robustness is critical to building trustworthy and resilient autonomous aerial systems for diverse application domains.

\section{Potential Solutions and Research Opportunities}
\subsection{Hardware Innovations and Platform Scalability}

To overcome the performance and deployment limitations of current agentic UAV systems, significant innovation in hardware design and platform scalability is essential. While much of the progress in autonomy has focused on software intelligence, foundational advances in aerial robotics hardware are required to ensure these systems can perform reliably in complex environments, carry diverse sensing payloads, and operate over extended missions. Emerging research in hybrid configurations, modular architecture, power optimization, and swarm scalability is driving the next generation of robust, adaptable agentic UAV platforms.

One of the most promising avenues is the adoption of \textbf{Vertical Take-Off and Landing (VTOL)} platforms \cite{gu2017development, misra2022review, qi2025model}. VTOL UAVs combine the agility of rotary-wing aircraft with the endurance of fixed-wing designs, enabling both hovering for high-resolution data capture and efficient forward flight for long-range missions. For example, fixed-wing VTOLs are particularly suited for surveillance over large agricultural fields or coastlines, while offering precise hovering over points of interest. This hybrid maneuverability supports more complex agentic behaviors such as waypoint replanning, terrain-following, and adaptive loitering all essential for autonomous aerial intelligence.

\textbf{Modular payload architectures} enable UAVs to adapt to a wide range of missions by dynamically reconfiguring their sensor and actuator configurations \cite{saponi2022embedded, kotarski2021modular}. Research into hot-swappable payload bays, plug-and-play sensor interfaces \cite{li2021plug, cui2024fastsim}, and autonomous self-recognition protocols allows agentic UAVs to autonomously determine available sensing capabilities and recalibrate their mission plans accordingly. For instance, a UAV equipped for environmental monitoring can automatically switch from thermal imaging to multispectral sensing depending on the observed ecosystem, or eject a seeding payload and attach a camera module for post-seeding assessment all without human intervention.

In the realm of endurance, \textbf{advanced power systems} are a critical enabler \cite{rajabi2023drone}. Innovations in high-energy-density batteries, supercapacitors for burst power, and lightweight fuel-cell technologies offer longer flight durations without increasing platform weight. Solar-assisted UAVs, which harvest energy through onboard photovoltaic panels, present a promising solution for high-altitude, persistent monitoring tasks. Moreover, wireless charging stations and dynamic recharging drones are being tested for energy-aware swarming operations, where a portion of the UAV fleet autonomously rotates in and out of operation while others recharge.

\textbf{Hybrid UAV designs}, combining multi-modal mobility such as aerial-terrestrial \cite{zhang2024development, tang2025duawlfin} or aerial-aquatic capabilities \cite{wu2024design, yao2023review}, represent a new frontier in platform versatility \cite{ducard2021review}. For example, UAVs that can land on water bodies or traverse short terrestrial paths can conduct inspections of infrastructure spanning multiple domains bridges, dams, and wetlands. In ecological surveillance, amphibious UAVs equipped with floating sensors or water samplers allow for richer multi-domain environmental data collection, extending the operational scope of agentic intelligence.

A major enabler for large-scale deployment is \textbf{swarm-based scalability}. Agentic swarms, composed of dozens or hundreds of UAVs with decentralized reasoning and coordination, can dynamically distribute tasks based on energy levels, spatial distribution, or sensing specialization. Swarm formations optimize mission efficiency by covering large areas, creating dynamic mesh networks, and recovering from individual drone failures without interrupting the global objective. Hardware innovations supporting these swarms include lightweight inter-UAV communication modules, onboard relative localization units (e.g., UWB, visual odometry), and collision-avoidance sensors optimized for close-proximity flight.

\textbf{Thermal management and environmental robustness} are also key hardware concerns in real-world deployment \cite{yang2024methods, li2021study}. Compact onboard computers used in agentic UAVs generate significant heat, which can degrade performance or cause system failure. Novel materials and passive cooling systems such as graphene coatings, vapor chambers, or aerodynamic airflow ducts are being explored to dissipate heat efficiently. Additionally, weather-proofing technologies such as conformal coating, anti-condensation systems, and vibration-resistant enclosures extend UAV survivability in rain, snow, and high-wind conditions.

In conclusion, advancing agentic UAV hardware demands a systemic redesign of platforms to support modularity, longevity, adaptability, and scalability. By integrating hybrid mobility, smart power systems, plug-and-play sensing, and swarming capabilities, the next generation of agentic UAVs will be capable of executing extended, complex missions across diverse environments. These innovations form a critical bridge between high-level AI capabilities and real-world operational reliability.

\subsection{Advances in Learning and Decision Systems}

At the heart of Agentic UAVs lies the capacity to learn from experience, reason under uncertainty, and make decisions autonomously across varying missions and environments. However, the challenges of generalization, semantic grounding, real-time adaptation, and safe decision-making have limited the deployment of intelligent aerial systems in real-world conditions. Recent advances in reinforcement learning \cite{bai2023toward}, federated learning \cite{nguyen2024flora},  VLMs , and memory-based reasoning \cite{singla2019memory, zhang2024multiple} are enabling more robust, explainable, and scalable decision architectures for UAV autonomy. These developments offer promising solutions to many of the core challenges outlined in earlier sections.

\textbf{Reinforcement learning (RL)} has emerged as a powerful paradigm for enabling autonomous UAVs to learn action policies through interaction with the environment \cite{koch2019reinforcement, liu2019reinforcement}. For tasks such as obstacle avoidance, terrain navigation, or target tracking, RL enables UAVs to optimize long-horizon performance metrics through trial and error \cite{garg2025reinforcement}. More recently, hierarchical RL and curriculum learning have been applied to teach UAVs complex behaviors \cite{yang2025hierarchical, mao2024dl} e.g., prioritizing mission goals, adjusting flight behavior based on weather or payload, and responding to emergency contingencies. In agentic settings, multi-agent reinforcement learning (MARL) allows UAVs in a swarm to coordinate effectively, optimize distributed coverage, and share information to achieve global objectives.

\textbf{Federated learning (FL)} addresses the challenge of data scarcity, privacy, and model robustness in distributed UAV deployments \cite{zhang2025latency, al2025privacy}. Instead of transferring sensitive aerial data to a central server, each UAV trains a local model using its onboard data and only shares model updates with a global aggregator \cite{alaya2025state}. This approach ensures data locality, reduces communication overhead, and allows the model to generalize across heterogeneous flight environments \cite{alsharif2024contemporary}. For example, agentic UAVs operating in different climate zones or terrains can contribute to a shared object detection model without compromising regional privacy or bandwidth. FL also enables continuous model refinement as new data is collected, supporting lifelong learning in UAV fleets.

\textbf{ VLMs } such as Flamingo \cite{alayrac2022flamingo}, OpenFlamingo \cite{awadalla2023openflamingo}, or GPT-4V \href{https://openai.com/index/gpt-4v-system-card/}{GPT-4V} have demonstrated strong capabilities in interpreting human instructions, generating semantic maps, and performing multimodal reasoning. By integrating VLMs into the decision pipeline of agentic UAVs, new capabilities emerge such as responding to natural language commands (e.g., “scan the east ridge for fallen trees”), interpreting scene context (e.g., “vehicles near the collapsed structure”), and asking clarification queries. VLMs enhance the semantic grounding of UAV decisions, linking visual perception to high-level goals. Their inclusion enables UAVs to perform adaptive mission planning, dynamically adjust objectives, and operate more intuitively in human-centric tasks such as disaster response or infrastructure auditing.

\textbf{Memory-augmented architectures} improve UAV decision-making by enabling the storage and recall of past experiences, environments, and action outcomes \cite{wang2022human, qiu2024memory}. Episodic memory modules allow UAVs to recognize previously visited locations, avoid repeating suboptimal strategies, or reuse successful plans in similar contexts. Combined with world models, these memory systems support planning over long horizons, enabling UAVs to anticipate future states and consequences of their actions. For example, in agricultural monitoring, UAVs can recall crop stress patterns from previous flights and prioritize areas with recurring issues.

\textbf{Task decomposition and modular learning frameworks} provide a scalable way to handle complex multi-stage missions \cite{choi2023modular}. Instead of learning monolithic end-to-end policies, agentic UAVs can decompose tasks into subtasks such as takeoff, navigation, inspection, data transmission and learn specialized policies for each module. These modules can then be composed into higher-level control graphs using behavior trees or neural program interpreters. This approach enhances interpretability, reusability, and debugging, which are critical for safety-critical deployments. For instance, in infrastructure inspection, separate modules may control flight stabilization, target alignment, and defect recognition, allowing for isolated testing and fine-tuning.

To further improve robustness, \textbf{uncertainty-aware decision systems} are being integrated \cite{shi2025collaborative, khatiri2025uncertainty}. Bayesian deep learning and ensemble models enable UAVs to estimate the confidence in their perception or control outputs. When uncertainty exceeds a threshold, the system can default to a safe action, request human intervention, or engage in re-exploration. This prevents overconfident errors in ambiguous conditions and builds trust in autonomous systems.

In conclusion, advances in learning and decision systems are foundational to realizing resilient, adaptable, and trustworthy agentic UAVs. By leveraging reinforcement learning, federated architectures, VLMs, and modular task reasoning, UAVs can achieve higher levels of autonomy and generalization, ultimately fulfilling their potential as intelligent agents in dynamic, real-world environments.

\subsection{Human-AI Interaction and Usability Enhancements}

As Agentic UAVs become more autonomous and complex in their decision-making, the need for intuitive, transparent, and trustworthy human-AI interaction becomes increasingly critical. Traditional UAV interfaces are often designed for expert operators, requiring manual waypoints, telemetry reading, and fine-grained control. However, Agentic UAVs capable of reasoning, adapting, and acting independently demand a paradigm shift toward user-centric design, natural communication modalities, and actionable explainability. Enhancing usability and human-AI collaboration will be vital to ensuring adoption, safety, and mission effectiveness in real-world deployment scenarios.

A key enabler of seamless interaction is the use of \textbf{natural language and voice command interfaces} \cite{savenko2024command}. Instead of relying on rigid command structures or predefined flight plans, users should be able to instruct UAVs using high-level semantic language such as “survey the south field for plant stress” or “inspect the rooftop AC unit and report damage.” Advances in  VLMs  have made it possible for UAVs to ground such instructions in real-time visual and spatial contexts, enabling agents to reason about tasks, recognize relevant objects, and plan actions accordingly. Voice-enabled command pipelines further reduce operational barriers, especially in field environments such as disaster response or construction zones where typing or manual control is impractical.

Another crucial area is \textbf{explainability and mission transparency} \cite{alharbi2023assuring, javaid2025explainable, banimelhem2023explainable}. As UAVs begin to make autonomous decisions such as choosing alternative routes, prioritizing tasks, or refusing unsafe actions users must understand the rationale behind these behaviors. This is particularly important in high-stakes domains like security surveillance, infrastructure inspection, and agriculture, where UAV behavior affects mission outcomes and trust. Agentic UAVs should be able to produce concise, interpretable explanations in real time, such as “Path adjusted due to wind gusts exceeding 25 km/h” or “Area skipped due to occlusion by canopy.” Explainable AI (XAI) techniques, such as saliency mapping, attention visualization, and decision summaries, are increasingly being integrated to allow users to audit perception outputs and control logic \cite{dissanayaka2024explainable, abdollahi2021urban}.

\textbf{User-centric interface design} is essential for democratizing UAV usage across diverse operator profiles from domain experts like farmers and engineers to public safety officers with limited technical training \cite{karjalainen2017social}. Interfaces should support multimodal interaction, combining map-based mission planning with visual overlays, natural language interaction, and alert summaries. Dashboards must be minimalistic yet informative, conveying UAV status, mission progress, remaining energy, and confidence scores. Adaptive interface strategies, which change based on user expertise or task complexity, are also being explored. For instance, expert users may prefer full autonomy with telemetry access, while novice users may require step-by-step decision visualization and validation prompts.

Another frontier is \textbf{shared autonomy and human-in-the-loop frameworks}, where human operators retain supervisory or collaborative roles in the UAV’s decision-making pipeline \cite{agrawal2021explaining, emami2024human}. In shared autonomy, UAVs propose plans or actions that can be approved, modified, or overridden by human users. This balances efficiency with control, especially in ambiguous or novel scenarios. For example, in wildlife conservation, a UAV may detect a potential animal cluster and seek operator validation before initiating a closer inspection to avoid disturbing the habitat. Human-on-the-loop systems are particularly useful in multi-agent swarm scenarios, where one operator may supervise several UAVs executing distributed tasks with partial autonomy.

To promote safety, \textbf{fail-safe behavior articulation} must also be improved \cite{siewert2019fail}. When faced with uncertainty, component failure, or adversarial conditions, the UAV should not only initiate safe maneuvers such as hovering, returning to home, or soft-landing but also clearly communicate the reason and action taken. Integrating behavioral transparency into emergency protocols ensures that users maintain confidence in autonomous operation, even when the UAV adapts to unanticipated conditions.

Lastly, \textbf{context-aware user interaction} enhances relevance and reduces cognitive load \cite{wu2025context}. Agentic UAVs can tailor alerts, reports, and requests based on situational context e.g., during forest fire monitoring, visualizing fire spread prediction maps or prioritizing alerts from high-risk zones. Such context-sensitive behaviors, combined with proactive recommendations (“Would you like to launch a follow-up survey?”), allow for a more intuitive and mission-aligned user experience.

In conclusion, advancing human-AI interaction and usability in Agentic UAVs is essential for practical deployment and public trust. By integrating voice commands, explainable models, shared autonomy, and user-adaptive interfaces, UAVs can operate not only as autonomous agents but also as collaborative partners bridging the gap between machine intelligence and human intent.

\section{Future Roadmap}
\subsection{Toward Fully Agentic Aerial Ecosystems}

As autonomous aerial systems continue to mature, the future of Agentic UAVs will evolve beyond isolated, task-specific deployments into distributed, self-governing ecosystems capable of persistent collaboration, adaptation, and self-improvement. The concept of a fully agentic aerial ecosystem envisions UAVs not merely as tools, but as intelligent agents participating in reflective decision-making, long-term coordination, and sustainable autonomy across domains. Achieving this vision necessitates progress along three key dimensions: self-evolving intelligence, reflective control architectures, and integration across the autonomy continuum.

First, the shift toward \textbf{self-evolving agents} will be central. Current UAVs, while capable of learning and adapting within predefined constraints, lack the capacity for autonomous skill acquisition, continual task refinement, and goal reinterpretation. Future agentic UAVs will incorporate lifelong learning architectures capable of acquiring new skills from raw experience, adapting to changing environmental contexts, and refining their mission policies through recursive evaluation. Such systems will leverage advancements in meta-learning, world modeling, and autonomous curriculum generation to optimize performance across dynamic mission spaces without human retraining.

Second, the development of \textbf{reflective control systems} will enhance UAV self-awareness and adaptability. Reflective control enables an agent to reason not only about the external world but also about its own goals, actions, and limitations. Through multi-level policy introspection, UAVs will be able to detect when their current strategy is inadequate, diagnose the source of failure, and reconfigure their planning modules in real time. This capability will be critical in high-risk applications such as emergency response, defense, and cooperative swarm operations, where environmental and mission uncertainty are high and static control policies may fail.

Finally, building \textbf{interconnected autonomy continua} across heterogeneous agents will unlock cooperative intelligence. Rather than relying solely on isolated UAVs, future ecosystems will feature heterogeneous fleets operating across aerial, terrestrial, and aquatic domains with fluid task allocation and shared semantic representations. These agents will dynamically transition along the autonomy continuum from supervised to fully autonomous based on mission criticality, operator availability, and real-time uncertainty. Aerial swarms will interoperate with satellite systems, ground robots, and IoT networks, forming intelligent environmental overlays capable of large-scale perception, decision, and action.

In summary, the future of Agentic UAVs lies in the convergence of self-evolving intelligence, reflective planning, and ecosystem-level coordination laying the foundation for fully autonomous aerial ecosystems that can adapt, learn, and act responsibly in complex, unstructured environments.

\subsection{System Integration and Collaborative Intelligence}

The future of agentic UAVs hinges not only on the advancement of individual autonomy but also on their seamless integration into broader, cooperative intelligent systems. Moving forward, the emphasis will shift from stand-alone UAV platforms to interconnected agents embedded within a shared ecosystem of machines, sensors, digital infrastructures, and human stakeholders. This transition toward \textbf{collaborative intelligence} necessitates innovations in interoperability, cross-domain coordination, and real-time decision fusion across the air-ground-cloud continuum.

A critical direction is the development of robust \textbf{UAV-UAV collaboration frameworks}. Future agentic UAVs will operate in dynamic multi-agent teams, capable of task delegation, behavior synchronization, and emergent coordination without centralized control. Swarm architectures will leverage decentralized communication and onboard reasoning to perform tasks such as collaborative mapping, adaptive coverage, and distributed sensing. For instance, in disaster response, one UAV can lead terrain scouting while others simultaneously perform victim localization or supply drops based on shared mission state.

To enable these collaborative behaviors, \textbf{Vehicle-to-Everything (V2X)} communication will become foundational. V2X encompasses UAV-to-UAV (V2V), UAV-to-ground station (V2G), and UAV-to-infrastructure (V2I) protocols, enabling real-time data exchange, traffic coordination, and risk mitigation. V2X standards will be crucial for integrating UAVs into urban environments, where interactions with smart cities, air traffic control, and autonomous ground vehicles require strict synchronization.

\textbf{Internet of Things (IoT)} and \textbf{edge-cloud orchestration} will further facilitate real-time integration with environmental sensors, stationary robots, and smart assets. Agentic UAVs will act as mobile edge nodes that collect, fuse, and relay data across sensor networks. For instance, in precision farming, UAVs can communicate with soil sensors and irrigation controllers to adjust agricultural interventions based on real-time conditions.

\textbf{Digital twin technologies} will serve as virtual mirrors of real-world systems, enabling continuous synchronization between the physical UAV environment and its computational replica. Through UAV-enabled scanning and feedback, digital twins can update simulations in near real time, supporting predictive maintenance, operational optimization, and scenario testing. This integration will be critical in infrastructure inspection, construction, and environmental forecasting.

Finally, \textbf{air-ground integration} with autonomous vehicles and ground robots will support end-to-end autonomy in logistics, agriculture, and industrial automation. Agentic UAVs will collaborate with UGVs for coordinated object delivery, relay communication in terrain-constrained regions, or share semantic maps for cross-platform reasoning.

In sum, the future of agentic UAVs will be defined by their ability to interoperate, collaborate, and co-evolve within intelligent cyber-physical ecosystems marking a shift from autonomous aerial platforms to synergistic partners in the broader landscape of intelligent machines.

\subsection{Sustainability, Equity, and Societal Impacts}

As Agentic UAVs mature into versatile autonomous systems, it is critical to align their development with broader goals of environmental sustainability, social equity, and global resilience. The deployment of these technologies must not only advance autonomy and intelligence but also support ethical, inclusive, and regenerative futures. The future roadmap for Agentic UAVs must therefore consider their systemic role in promoting climate resilience, enabling equitable access, and amplifying citizen participation in digital and ecological governance.

A key direction is enhancing \textbf{accessibility for smallholder farmers and under-resourced communities}. While UAVs have revolutionized large-scale agriculture and logistics, their cost, complexity, and infrastructure requirements have largely excluded smallholders in developing regions. Future agentic UAVs must embrace affordability, simplicity, and offline capability. Modular low-cost platforms with intuitive voice-command interfaces and autonomous self-deployment will democratize access to aerial intelligence. Open-source agentic software stacks, pre-trained VLMs optimized for local dialects, and solar-powered UAVs are potential solutions that reduce dependency on connectivity and skilled labor. These developments will enable smallholders to monitor crop stress, detect pests, and optimize inputs, bridging digital divides and enhancing food security.

Another frontier is \textbf{resource-efficient and climate-resilient UAV operation}. The development of energy-aware flight planning, lightweight materials, and hybrid power systems will reduce the carbon footprint of UAV missions. Swarm coordination can be optimized for minimal overlap and battery usage, and UAVs can dynamically adapt flight behaviors based on solar irradiance, wind speed, or thermal updrafts to conserve energy. In climate-vulnerable regions, agentic UAVs can autonomously monitor early warning indicators such as rising water levels, crop drought stress, or wildfire outbreaks offering communities the ability to prepare and adapt in real time. Integration with environmental IoT sensors and regional climate models will enhance the precision and responsiveness of these systems.

\textbf{Participatory sensing and citizen engagement} represent transformative opportunities for agentic UAVs. Unlike traditional top-down deployments, future systems can be co-designed with local stakeholders who guide where, when, and how UAVs collect data. In urban areas, residents can request air quality measurements, map green infrastructure, or assess heat islands via natural language queries. In rural areas, local rangers can collaboratively plan anti-poaching patrols or biodiversity surveys. Through real-time explainability, mission transparency, and data-sharing dashboards, agentic UAVs can foster trust and inclusion while enhancing situational awareness for communities.

Furthermore, agentic UAVs can play a pivotal role in \textbf{environmental justice and equitable development}. By autonomously monitoring illegal logging, pollution hotspots, or infrastructure neglect in underserved regions, UAVs can generate transparent, verifiable evidence to support advocacy and policy change. These systems can help identify gaps in disaster recovery, track aid delivery, or ensure fair distribution of resources during emergencies. Coupled with decentralized data storage and cryptographic provenance tracking, UAV-collected data can enhance accountability and promote evidence-based decision-making at local and national levels.

Looking ahead, \textbf{institutional integration and cross-sector collaboration} will be vital to amplify societal impact. Public-private partnerships, participatory governance frameworks, and AI-for-Good consortia must shape how agentic UAVs are designed, deployed, and evaluated. Governments and NGOs can establish ethical guidelines, ensure compliance with data protection laws, and provide funding for UAV deployments in marginalized areas. Universities and local innovators can co-create agentic applications tailored to regional needs, from ecological restoration to urban planning.

In summary, the future of Agentic UAVs must be guided not only by technical sophistication but by ethical foresight, equity-centered design, and sustainability imperatives. By ensuring inclusive access, optimizing for environmental performance, and amplifying citizen engagement, agentic aerial systems can become powerful tools for achieving climate resilience, social justice, and digitally empowered societies.

\section{Conclusion}

This review has systematically explored the emergence and advancement of Agentic UAVs an evolution from traditional autonomous systems to intelligent aerial agents capable of perception, reasoning, planning, and learning across dynamic real-world environments. Beginning with foundational components and enabling technologies, we analyzed how these systems differ from conventional UAVs in architecture and cognitive capabilities. A comprehensive multi-domain synthesis then showcased how Agentic UAVs are being deployed in agriculture, disaster response, infrastructure inspection, logistics, defense, and ecological monitoring. We also identified major challenges related to hardware constraints, regulation, safety, and data reliability. Subsequently, potential solutions were proposed, including innovations in edge hardware, federated learning, explainable autonomy, and human-agent interfaces. Finally, the roadmap toward fully agentic aerial ecosystems emphasized sustainability, collaborative intelligence, and societal inclusiveness. This review serves as both a foundational resource and a strategic blueprint for researchers, developers, and policymakers aiming to build the next generation of intelligent aerial systems.

\begin{itemize}
  \item \textbf{Agentic UAVs represent a paradigm shift from automation to autonomy through cognitive architectural components.} Traditional UAVs follow pre-programmed flight paths and fixed sensor routines, whereas Agentic UAVs integrate perception, cognition, decision-making, and communication in real time. These systems rely on  VLMs , multimodal sensors, onboard reasoning, and memory-augmented architectures to dynamically perceive environments and generate mission-specific responses. Architectural elements such as reflective planning, state tracking, and action adaptation enable UAVs to operate under uncertainty and revise goals on the fly. The agentic stack supports semantic understanding, self-monitoring, and recursive behavior generation, moving UAV intelligence beyond waypoint navigation toward true aerial agency.

  \item \textbf{Agentic UAVs are rapidly extending into diverse domains, showcasing transformative use cases with high societal and operational impact.} In precision agriculture, they autonomously monitor crop health and optimize seeding or spraying. In disaster response, they conduct search and rescue with adaptive terrain mapping. Environmental UAVs track biodiversity and air quality, while those in infrastructure inspect bridges or towers using AI-based damage detection. In logistics, swarm-enabled delivery systems support last-mile fulfillment. Defense UAVs perform autonomous patrols and threat detection, and in conservation, UAVs map habitats and detect poaching events. These domain-specific deployments illustrate the generalizability, responsiveness, and mission awareness that distinguish agentic systems from their predecessors.

  \item \textbf{Despite their promise, Agentic UAVs face key challenges across technical, regulatory, and cognitive dimensions.} Operationally, they are limited by short battery life, payload capacity, and compute constraints that hinder long-range missions and real-time processing. Navigational safety in GPS-denied or cluttered environments requires reliable sensor fusion and fallback protocols. Regulatory frameworks struggle to accommodate BVLOS operations, data privacy, and ethical surveillance, creating uncertainty for deployment. From a cognitive perspective, generalization remains limited models may fail in out-of-distribution environments, and semantic errors in perception can cascade into unsafe decisions. Transparency and trust are also hampered by opaque black-box models, especially in safety-critical settings.

  \item \textbf{Recent solutions point toward scalable, explainable, and adaptive UAV architectures enabled by hardware and algorithmic innovation.} On the hardware front, hybrid VTOL designs, modular payloads, and power-optimized UAVs extend operational flexibility. Swarm coordination enhances spatial coverage and resilience. On the intelligence side, reinforcement learning and federated learning support context-aware behavior and decentralized knowledge sharing. Vision-language models offer semantic grounding, while memory-based reasoning enables task continuity. Explainable interfaces, human-in-the-loop control, and intuitive UI designs are being integrated to bridge AI decision-making with user understanding, allowing collaborative mission design, safe overrides, and post-hoc explanations for mission behavior.

  \item \textbf{The future of Agentic UAVs lies in fully integrated, ecosystem-level aerial intelligence with equity, sustainability, and human participation at its core.} UAVs will form cooperative networks with ground robots, IoT devices, and cloud-based digital twins, supporting persistent autonomy and multi-agent collaboration. Self-evolving agents will continuously adapt through lifelong learning and reflective control. Equity-focused innovations such as low-cost, voice-interactive UAVs will enable adoption by smallholders and marginalized communities. Sustainability will be addressed via energy-aware path planning and UAV-assisted environmental sensing. Through participatory AI and community-driven deployment, Agentic UAVs can become instruments of resilience, justice, and inclusive innovation across geographies and sectors.
\end{itemize}

\bibliographystyle{unsrt}  
\bibliography{references}






\end{document}